\documentclass{article}
\PassOptionsToPackage{numbers, compress}{natbib}

\usepackage[implicit=false]{hyperref}
\usepackage[preprint]{corl_2022} 

\usepackage{amsmath,amssymb,graphicx}
\usepackage{wrapfig}
\usepackage{multirow}
\usepackage{adjustbox}
\usepackage{enumitem}
\usepackage{booktabs}
\usepackage{xspace}
\usepackage{xcolor}
\usepackage[noblocks]{authblk}

\usepackage[english]{babel}
\newtheorem{theorem}{Theorem}[section]

\newtheorem{lemma}[theorem]{Lemma}

\newcommand{\attack}{\textit{Deterministic Attack}\xspace}
\newcommand{\context}{\textit{Context Attack}\xspace}
\newcommand{\latent}{\textit{Latent Attack}\xspace}
\newcommand{\method}{\textit{RobustTraj}\xspace}

\usepackage{amsmath,amsfonts,bm}

\def\eqref#1{equation~\ref{#1}}

\def\1{\bm{1}}

\DeclareMathOperator*{\argmax}{arg\,max}
\DeclareMathOperator*{\argmin}{arg\,min}

\title{Robust Trajectory Prediction\\ against Adversarial Attacks}

\author[1,2]{Yulong Cao\thanks{Work done during an internship at NVIDIA}\textsuperscript{\enspace}}
\author[2]{Danfei Xu}
\author[2]{Xinshuo Weng}
\author[1]{Z. Morley Mao}
\author[2,3]{Anima Anandkuma}
\author[2,4]{Chaowei Xiao}
\author[2,5]{Marco Pavone}
\affil[1]{University of Michigan}
\affil[2]{NVIDIA}
\affil[3]{California Institute of Technology}
\affil[4]{ASU}
\affil[5]{Stanford University}

\begin{document}
\maketitle


\begin{abstract}
  Trajectory prediction using deep neural networks (DNNs) is an essential component of autonomous driving (AD) systems.  However, these methods are vulnerable to adversarial attacks, leading to serious consequences such as collisions. In this work, we identify two key ingredients to defend trajectory prediction models against adversarial attacks including (1) designing effective adversarial training methods and (2) adding domain-specific data augmentation to mitigate the performance degradation on clean data. We demonstrate that our method is able to improve the performance by 46\% on adversarial data and at the cost of only 3\% performance degradation on clean data, compared to the model trained with clean data. Additionally, compared to existing robust methods, our method can improve performance by 21\% on adversarial examples and 9\% on clean data. Our robust model is evaluated with a planner to study its downstream impacts. We demonstrate that our model can significantly reduce the severe accident rates (e.g., collisions and off-road driving)\footnote[1]{Our project website is at \href{https://robustav.github.io/RobustTraj}{https://robustav.github.io/RobustTraj}}.
\end{abstract}

\section{Introduction}
Trajectory prediction is a critical component of modern autonomous driving (AD) systems. It allows an AD system to anticipate the future behaviors of other nearby road participants and plan its actions accordingly.
Recent trajectory prediction models built on Deep Neural Networks (DNN) have demonstrated state-of-the-art performance on large-scale benchmarks~\cite{alahi2016social,ivanovic2019trajectron,salzmann2020trajectron++,yuan2021agent,rhinehart2018r2p2,rhinehart2019precog,kosaraju2019social},
showing a promising path towards learning-based trajectory prediction for AD systems. 
As trajectory prediction plays an important role in AD systems, accurate predictions are required for making safe driving decisions. It is crucial to understand how unknown scenarios will affect trajectory predictions and then bolster the robustness of such trajectory predictions in return. 

\begin{figure}
    \centering
    \small
    \includegraphics[width=\linewidth]{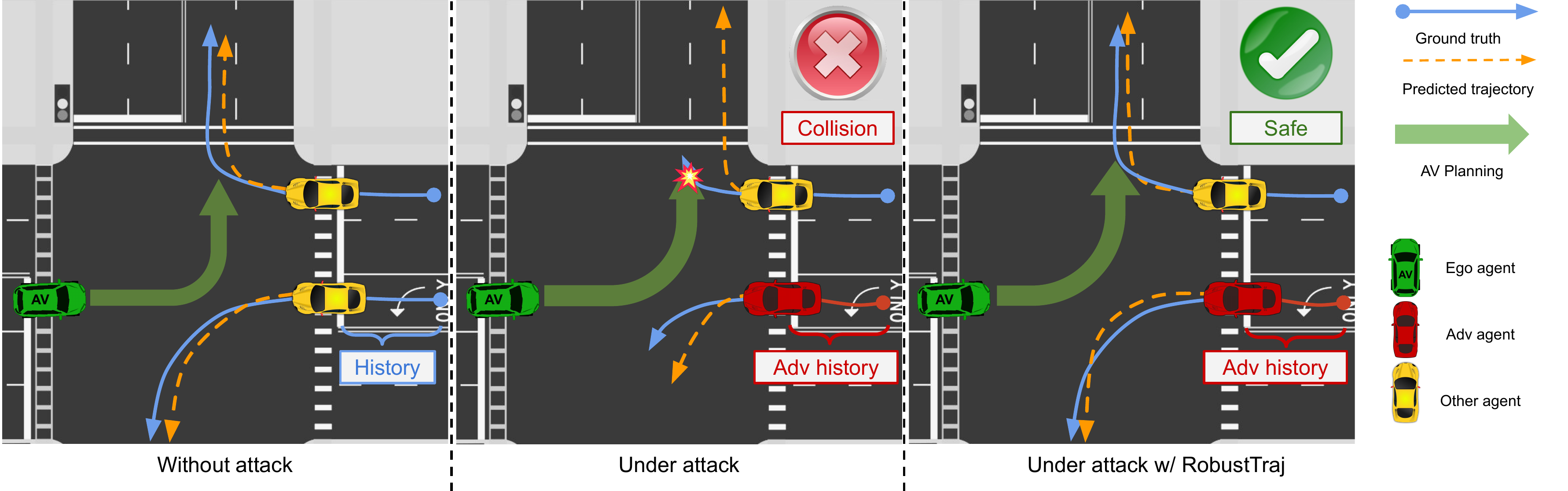}
    \caption{Overview of \method preventing Autonomous Vehicle (AV) from collisions when its trajectory prediction model is under adversarial attacks. When the trajectory prediction model is under attack, the AV predicts the wrong future trajectory of the other agent turning right (yellow vehicle). This results in AV speeding up instead of slowing down, and eventually colliding into the other vehicle.
    }
    \label{fig:overview}
\end{figure}

 To achieve this goal, adversarial attacks~\cite{madry2017towards,szegedy2013intriguing,carlini2017towards} are often used as a proxy to measure the worst-case performance of the model when facing unseen scenarios. Similarly, we use a standard adversarial attack setup~\cite{zhang2022adversarial} for trajectory predictions.
As illustrated in Fig.~\ref{fig:overview},
an adversarial agent (red vehicle) aims to cause a traffic accident. It drives along a carefully designed trajectory (i.e., adv history) to influence the trajectory prediction model of the Autonomous Vehicle (green vehicle).
Such an adversary can critically compromise the predicted trajectories of all other agents  by altering its route in inconspicuous ways.  
By fooling the trajectory prediction models, it can further affect downstream planning of the AV systems and cause serious consequences.
Using the adversarial attack as the proxy, this work aims to develop effective techniques to bolster the robustness of trajectory prediction models against adversarial attacks and improve the AD's safety under uncertain scenarios.


At the same time, adversarial robustness for machine learning is a widely-studied area,
but most works focus on classification tasks~\cite{defense:jeddi2020simple,defense:liu2019extending,defense:papernot2016distillation,defense:papernot2017extending,defense:shafahi2019adversarial,defense:wong2020fast,defense:xie2020adversarial,defense:Xie2020Intriguing,defense:xie2020smooth,defense:xu2017feature,defense:yang2019me,defense:zhang2019theoretically}. 
Among the proposed techniques, adversarial training~\cite{madry2017towards} remains the most effective and widely used method to defend classifiers against adversarial attacks. The general strategy of adversarial training is to solve a min-max game by generating adversarial examples for a model at each training step and then optimizing the model to make correct predictions for these samples. However, directly applying adversarial training to trajectory prediction presents a number of critical technical challenges. 

First, 
most trajectory prediction methods employ probabilistic generative models to cope with the uncertainty in motion forecasting~\cite{ivanovic2019trajectron,salzmann2020trajectron++,yuan2021agent,rhinehart2018r2p2,rhinehart2019precog,kosaraju2019social}. As we will show in this paper, the stochastic components of these models (e.g., posterior sampling in VAEs) can obfuscate the gradients that guide the adversarial generation, making na\"ive adversarial training methods ineffective. 
Second, adversarial training on trajectory prediction task aims to model joint data distribution of future trajectories and adversarial past trajectories. However, the co-evolution of the adversarial sample distribution and the prediction model during the training process makes the joint distribution hard to model and often destabilizes the adversarial training.
Finally, prior work~\cite{defense:zhang2019theoretically} shows that adversarial training often leads to degraded performance on clean (unperturbed) data, while retaining good performance in benign cases is crucial due to the critical role of trajectory prediction for AVs. Hence, an effective adversarial training method must carefully balance the benign and the adversarial performance of a model.


\textbf{Our approach.} We propose an adversarial training framework for trajectory predictions named \method, by addressing the aforementioned challenges. 
First, to address the issue of an obfuscated gradient in adversarial generation due to stochastic components, we devise a \textit{deterministic attack} that creates a deterministic gradient path within a probabilistic model to generate adversarial samples. 
Second, to address the challenge of an unstable training process due to shift in adversarial distributions,
we introduce a hybrid objective that interleaves the adversarial training and learning from clean data to anchor the model output on stable clean data distribution.
Finally, to achieve balanced performances on both adversarial and clean data, we introduce a domain-specific data augmentation technique for trajectory prediction via a dynamic model. This data augmentation technique generates diverse, realistic, and dynamically-feasible samples for training and achieves a better performance trade-off on clean and adversarial data.


We empirically show that \method can effectively defend two different types of probabilistic trajectory prediction models~\cite{yuan2021agent,kosaraju2019social} against adversarial attacks, while incurring minimal performance degradation on clean data. 
For instance, \method can increase the adversarial performance of AgentFormer~\cite{yuan2021agent}, a state-of-the-art trajectory prediction model, by 46\% at the cost of 3\% performance drop on clean data.
To further show impacts of our method on the AD stack, we plug our robust trajectory prediction model into a planner and demonstrate that our model reduces  serious accidents rates (e.g., collisions and off-road driving) under attacks by 100\%, compared to the standard non-robust model trained using only clean data.

\section{Related Work}
\label{sec:related_work}



\noindent\textbf{Adversarial attacks and defenses on trajectory prediction.}
\label{subsec:related_work_adv_attacks_defenses}
A recent work began to study the adversarial robustness of trajectory prediction models~\cite{zhang2022adversarial}. \citet{zhang2022adversarial} demonstrated that perturbing agents' observed trajectory can adversarially impact the prediction accuracy of a DNN-based trajectory forecasting model. To mitigate the issue, \citet{zhang2022adversarial} proposed several defense methods such as data augmentation and trajectory smoothing.
However, these methods are less effective when facing adaptive attacks~\cite{athalye2018obfuscated}. 
In our work, we propose to use adversarial training which provides the general adversarial robustness that can resist adaptive attacks.

\noindent\textbf{Adversarial training.}
\label{subsec:related_work_adv_training}
 A variety of adversarial training methods have been proposed to defend DNN-based models against adversarial attacks~\cite{madry2017towards,defense:Xie2020Intriguing,defense:jeddi2020simple,defense:liu2019extending,defense:papernot2016distillation,defense:papernot2017extending,defense:shafahi2019adversarial,defense:wong2020fast,defense:xie2020adversarial,defense:Xie2020Intriguing,defense:xie2020smooth,defense:xu2017feature,defense:yang2019me,defense:zhang2019theoretically}. The most common strategy is to design a min-max game with the inner maximization process  and outer minimization process. The inner maximization process generates adversarial examples that maximize an adversarial objective (e.g., make wrong prediction). The outer minimization process then updates the model parameters to minimize the error on the adversarial examples.  
Although there exists a large body of literature in studying adversarial robustness for machine learning, most focus on the problem of discriminative model (e.g., object recognition), leaving other problem domains (e.g., conditional generative models) largely unexplored. 
In this work, we develop a novel adversarial training method for trajectory prediction models, where most state-of-the-art trajectory prediction models are generative and probabilistic, by addressing a number of critical technical challenges. 

\section{Preliminaries and Formulation}\label{sec:pre}

\textbf{Trajectory prediction.} The goal is to predict future trajectory distribution $\mathcal{P}_\theta (Y | X)$  of $N$ agents conditioned on their $H$ history time states $\textbf{X} = \left( \textbf{X}^{-H+1}, \dotsc, \textbf{X}^0\right)$, and other environment context such as maps~\footnote{For simplify, we ignore contextual information.} to predict $T$ time-step future trajectories    $\textbf{Y} = \left( \textbf{Y}^1,\dotsc, \textbf{Y}^T \right)$ . 
For observed time steps $t \leq 0$, we denote the agent states as $\textbf{X}^t = \left( x_1^t,\dotsc, x_i^t, \dotsc, x_N^t \right)$, where $x_i^t$ is the state of agent $i$ at the time step $t$.
Similarly, $\textbf{Y}^t = \left(y_1^t,\dotsc,y_N^t\right)$ denotes the states of $N$ agents at a future time step $t$ ($t>0$). We denote the ground truth and the predicted future trajectories as $\textbf{Y}$ and $\hat{\textbf{Y}}$, respectively.  
We denote the history information encoded by a function $f$ as the decision context $\mathbf{C}=f(\mathbf{X})$.
\textbf{Probabilistic trajectory prediction models.} In this work, we focus on defending generative, probabilistic trajectory prediction models, as they have demonstrated superior performance in modeling uncertainty in predicting future motions~\cite{ivanovic2019trajectron,salzmann2020trajectron++,yuan2021agent,rhinehart2018r2p2,rhinehart2019precog,kosaraju2019social}. We consider two most popular types of generative models: conditional variational encoders (CVAEs) and conditional GANs (cGANs), both can be viewed as latent variable models. We define latent variables $\mathbf{Z} = \{z_1, ..., z_i, ..., z_N \}$ where $z_i $
represents the latent variable of the agent $i$. CVAE formulates the generative problem as:
$p_\theta (\mathbf{Y} | \mathbf{X})  = \int p_\theta (\mathbf{Y} | \mathbf{X}, \mathbf{Z}) \cdot  p_\theta (\mathbf{Z} | \mathbf{X}) d \mathbf{Z} $, where $p_\theta (\mathbf{Z} | \mathbf{X})$ is a conditional Gaussian prior ($ \mathcal{N}(p_\theta^\mu(\mathbf{Z}|\mathbf{X}),p_\theta^\sigma(\mathbf{Z}|\mathbf{X}))$) with mean $p_\theta^\mu(\mathbf{Z}|\mathbf{X})$ and standard deviation $p_\theta^\sigma(\mathbf{Z}|\mathbf{X})$; $p_\theta (\mathbf{Y} |\mathbf{X}, \mathbf{Z})$ is a conditional likelihood model.
The model is usually trained through optimizing a negative evidence lower objective~\cite{yuan2021agent}: 
\begin{equation} \label{eq:cvae_loss}
\begin{split}
\mathcal{L}_{\text{total}} &= \mathcal{L}_\text{elbo} + \mathcal{L}_{\text{diversity}} \\
&= - \mathbb{E}_{q_\phi(\mathbf{Z} | \mathbf{Y},\mathbf{X})} 	[\, \log p_\theta (\, \mathbf{Y} | \mathbf{Z}, \mathbf{X}\,) \,] 
 + \text{KL}(\, q_\phi(\mathbf{Z} | \mathbf{Y},\mathbf{X}) \parallel p_\theta (\, \mathbf{Z} |\mathbf{X}\,) \,)  + \min_k\parallel\hat{\mathbf{Y}}^{(k)}-\mathbf{Y}\parallel^2 \text{,}
\end{split}
\end{equation}

where $q_\phi(\mathbf{Z}|\mathbf{Y}, \mathbf{X})$ is an approximate posterior 
parameterized by $\phi$, $p_\theta(\mathbf{Z}|\mathbf{X})$ is a conditional Gaussian prior parameterized by $\theta$, and $p_\theta(\mathbf{Y}|\mathbf{Z}, \mathbf{X})$ is a conditional likelihood modeling future trajectory $\mathbf{Y}$  via the latent codes $\mathbf{Z}$ and past trajectory $\mathbf{X}$. 
Additionally, $\mathcal{L}_\text{diversity} = \min_k\parallel\hat{\mathbf{Y}}^{(k)}-\mathbf{Y}\parallel^2$ is a diversity loss, which encourages the
network to produce diverse samples. Given each past trajectory $X$, the model generates $K$ sets of latent codes $\{\mathbf{Z}^{(1)},\cdots, \mathbf{Z}^{(k)},\cdots, \mathbf{Z}^{(K)} \}$ from the conditional Gaussian prior $ \mathcal{N}(p_\theta^\mu(\mathbf{Z}|\mathbf{X}),p_\theta^\sigma(\mathbf{Z}|\mathbf{X}))$, where $\mathbf{Z}^{(k)} =\{z_1^{k}, \cdots,z_n^{k} \} $ , resulting in $K$ future trajectories $\hat{\mathbf{Y}}^{(k)}$. 

Similarly, in a conditional Generative Adversarial Net (cGAN)-based model (e.g., Social-GAN~\cite{alahi2016social}),
it uses a loss function as follows:
\begin{equation} \label{eq:cgan_loss}
\begin{split}
\mathcal{L}_\text{total} &= \mathcal{L}_\text{gan} + \mathcal{L}_\text{diversity} \\
&= \mathbb{E}_{\mathbf{Y}\sim p_\text{data}}[\,\log D_{\theta}(\mathbf{Y|X})\,] + 
\mathbb{E}_{\mathbf{Z}\sim p_\text{Z}} [\,\log(1-D_\theta(G_\phi(\mathbf{Y|X, Z})))\,]+ \min_k\parallel\hat{\mathbf{Y}}^{(k)}-\mathbf{Y}\parallel^2 \text{, }
\end{split}
\end{equation}

where $G$ represents the generator and $D$ represents the discriminator. $\hat{\mathbf{Y}}^{(k)} = G(\mathbf{Y}|\mathbf{X},\mathbf{Z}^{(k)})$ is one of the predicted trajectories in which $\mathbf{Z}^{(k)}$ is randomly sampled from  $\mathcal{N}(0, 1)$.  

\textbf{Threat model.} 
We follow the setup in prior work ~\cite{zhang2022adversarial} and adopt an idealized threat model, where the adversary alters its observed history $\textbf{X}$ by adding a perturbation $\delta$ bounded by the adversarial set  $\mathbb{S}_p^\epsilon = \{\delta | \parallel\delta\parallel_p \leq \epsilon\}$, where $\epsilon$ is the perturbation bound. The perturbation aims to mislead the prediction $\hat{\textbf{Y}}$.
A na\"ive adversarial attack is to find the perturbation through an adversarial objective
$\delta = \argmax_{\delta  \in \mathbb{S}} \{ \min_{k \in \{1, \cdots, K\}}\parallel p_\theta(\mathbf{Y|X + \delta}, \mathbf{Z}^{(k)})-\mathbf{Y}\parallel^2 \}$, where $ p_\theta(\mathbf{Y|X + \delta}, \mathbf{Z}^{(k)})$ is the predicted trajectory conditioned on the  random variable $\mathbf{Z}^{(k)}$ and adversarial history trajectory $\mathbf{X}+\delta$.
\textbf{Na\"ive adversarial training.} Adversarial training formulates a min-max game with an inner maximization process that optimizes the perturbation $\delta$ to generate adversarial examples for misleading the model at each training iteration, and an outer minimization process that optimizes the model parameters to make correct predictions for these examples. 
We follow the standard adversarial training formulation~\cite{madry2017towards}:
\begin{equation}
\label{eq:adv_train}
    \min_{\theta,\phi} \,\max_{\delta \in \mathbb{S}}\quad \mathcal{L}_\text{total}(\,\mathbf{X}+\delta, \mathbf{Y}) \text{.}
\end{equation}

\section{\method{}: Robust Trajectory Prediction}
As stated earlier, applying adversarial training for trajectory prediction presents three critical challenges:
(1) gradient obfuscation due to model stochasticity, (2) unstable learning due to changing adversarial distribution, and (3) performance loss in the benign situation. In this section, we describe each challenge in more detail and present the corresponding solutions in our \method{} method.
\textbf{Improve adversarial generation with \attack.}  Since trajectory prediction is inherently uncertain and there is no single correct answer, probabilistic generative models are usually used to cope with the stochastic nature of the trajectory prediction task. Such stochasticity will obfuscate the gradients that are used to generate effective adversarial examples in the inner maximization process of adversarial training. 
The na\"ive attack  mentioned in section~\ref{sec:pre} is a straightforward way to achieve this goal. However, this optimization involves a stochastic sampling process $\mathbf{Z}^{(k)} \sim \mathcal{N}(p_\theta^\mu(\mathbf{Z}|\mathbf{X}),\, p_\theta^\sigma(\mathbf{Z}|\mathbf{X}))$. 
Such a stochastic process will obfuscate the gradients for finding the optimal adversarial perturbation $\delta$, making the outer minimization (robust training) less effective. 
In order to sidestep such stochasticity, we propose the \textit{deterministic attack} that creates a deterministic gradient path within the model to  generate the adversarial perturbation.
$\hat{\mathbf{Z}}$. Specifically, we use a {\em deterministic latent code} by replacing the sampling process $\mathbf{Z}^{(k)} \sim \mathcal{N}(p_\theta^\mu(\mathbf{Z}|\mathbf{X}),\,p_\theta^\sigma(\mathbf{Z}|\mathbf{X}))$, with the maximum-likelihood sample (here, i.e $\hat{\mathbf{Z}} = p_\theta^\mu(\mathbf{Z}|\mathbf{X})$). The objective for generating the adversarial perturbation is thus:
\begin{equation}
\label{eq:adv_loss}
    \delta = \argmax_{\delta \in \mathbb{S}} \mathcal{L}_\text{adv} (\mathbf{X} + \delta, \mathbf{Y}) = \, \argmax_{\delta \in \mathbb{S}} \parallel p_\theta(\mathbf{Y}|\hat{\mathbf{Z}},\mathbf{X+\delta}) -\mathbf{Y}\parallel^2 \text{, where } \hat{\mathbf{Z}} = p_\theta^\mu(\mathbf{Z}|\mathbf{X}+\delta) \text{.}
\end{equation}
We empirically show that gradients from this deterministic gradient path can effectively guide the generation of adversarial examples. We name our attack as \attack.

\textbf{Stabilize adversarial training with bounded noise and hybrid objective.}
During the adversarial training process,  the distribution of the perturbed input $\mathbf{X}+\delta$ coevolves with the training process as $\delta$ is calculated via an inner maximization process at each training iteration. 
Although $\delta$ is bounded by the adversarial set $\mathbb{S}_p^\epsilon$, the resulting latent condition variable $\mathbf{C}=f(\mathbf{X}+\delta)$ can be arbitrarily noisy since the Lipschitz constant of neural network layers ($f$) is not bounded during training (See \textit{Lemma 1.} in Appendix A). Since $\mathbf{C}=f(\mathbf{X}+\delta)$ is noisy, it is a less informative signal compared to the deterministic signal $X$.  
Thus, modeling $p_\theta(\mathbf{Y}|\mathbf{X}+\delta,\mathbf{Z})$  becomes substantially harder. In an extreme case that $\mathbf{C}=f(\mathbf{X}+\delta)$ is super noisy and contains no information, the training process can degenerate to model $p_\theta(\mathbf{Y}|\mathbf{Z})$, resulting in the undesirable worse performance on the clean data. 

\begin{figure}[t]
    \centering
    \begin{minipage}{.24\textwidth}
        \centering
        \includegraphics[width=\textwidth,trim={0 5.2cm 0 0},clip]{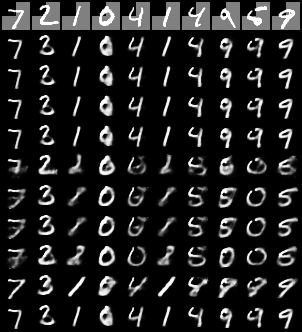}
        \text{(a) Clean}
    \end{minipage}
    \begin{minipage}{.24\textwidth}
        \centering
        \includegraphics[width=\textwidth,trim={0 5.2cm 0 0},clip]{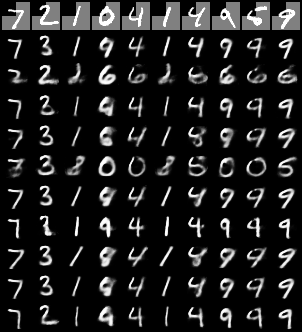}
        \text{(b) Salt and pepper}
    \end{minipage}
    \begin{minipage}{.24\textwidth}
        \centering
        \includegraphics[width=\textwidth,trim={0 5.2cm 0 0},clip]{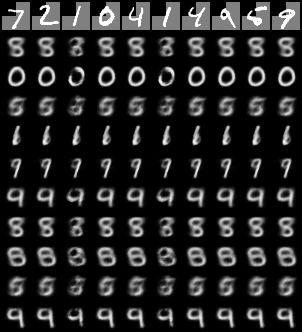}
        \text{(c) Adv noise}
    \end{minipage}
    \begin{minipage}{.24\textwidth}
        \centering
        \includegraphics[width=\textwidth,trim={0 0.2cm 0 1.5cm},clip]{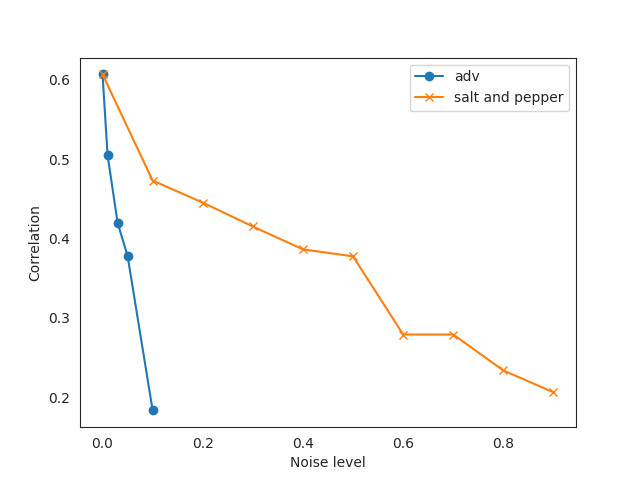}
        \text{(d) Correlation}
    \end{minipage}
    \caption{Visualizations of the CVAE models trained with clean (a) data, \textit{Salt and pepper} noise (b), and adversarial perturbations (c); Quantitative results of the correlation between the label of the generated images and conditioned images at different noise levels (d).
    }
    \vspace{-1em}
    \label{fig:cvae_analysis}
\end{figure}
\label{sec:noisy}
To further validate the above hypothesis that it is hard to model $p_\theta(\mathbf{Y}|\mathbf{X}+\delta,\mathbf{Z})$ with a changing data distribution of $\mathbf{X}+\delta$, we conduct an additional experiment. 
For simplicity, we use MNIST~\cite{deng2012mnist} as the dataset. 
As shown in Fig.~\ref{fig:cvae_analysis}, we divide each digit image into four quadrants. We take the top-left quadrant as the condition $\mathbf{X}$ and the remaining quadrants as the output $\mathbf{Y}$. We train a CVAE ($p_\theta(\mathbf{Y}|\mathbf{X}+\delta,\mathbf{Z})$) to model $\mathbf{Y}$ by using clean data ($\mathbf{X}$) or noisy data ($\mathbf{X}+\delta$), where $\delta$ represents salt and pepper noise~\cite{saltpepper-wiki:1085283109} or adversarial noise~\cite{madry2017towards}, resulting in Fig.~\ref{fig:cvae_analysis} (a), (b), (c) respectively. 
The top-left region of each image in the first row is the conditional variables $\mathbf{X}$. The rest of rows are the generated images with different $\mathbf{Z}$. Each column in the same row uses the same $\mathbf{Z}$. 
We observe that the model trained on clean data successfully captures the conditional distribution (i.e., the generated image highly depends on $\mathbf{X}$) while the model trained with adversarial noise degenerates and ignores the condition (i.e., each row generates images of the same digit). This result shows that the conditional generative model fails to learn from $\mathbf{X}$.
To provide a quantitative analysis, we measure the correlation between the label of the generated images  and the label of their conditioned image quadrants, resulting in Fig~\ref{fig:cvae_analysis} (d). More details on how to calculate the correlation are in the Appendix A.
We observe that the correlation drops as the noise level increases for both adversarial nose and \textit{salt and pepper} noise. 
Adversarial noise is more effective to degenerate the  conditional generative model. 
Therefore, we conclude that (1) the noises in the conditional data lead to degenerated conditional generative model (i.e., from CVAE to VAE); (2) the level of degeneration depends on the noise levels.

Based on the analysis result, to better learn a robust trajectory prediction model,
we need to bound $|f(\mathbf{X}+\delta) - f(\mathbf{X})|$ to reduce the noise level.
Hence, we propose the following regularization loss $\mathcal{L}_\text{reg}$:
\begin{equation}
\label{eq:reg}
\mathcal{L}_\text{reg} = \mathrm{d}( f(\mathbf{X}+\delta), f(\mathbf{X}))\text{, }
\end{equation}
where $\mathrm{d}$ is a distance function (e.g., we use $L_2$ norm as the distance metric).

In addition, because the clean data has a fixed distribution, simultaneously learning from the clean data during the adversarial training process anchors the conditional distribution on a stable clean data distribution.
Specifically, we propose the following hybrid objective:
\begin{equation}
\label{eq:clean}
\mathcal{L}_\text{clean}(\mathbf{X}, \mathbf{Y}) = \mathcal{L}_\text{total}(\mathbf{X},\mathbf{Y}) \text{, }
\end{equation}
where $\mathcal{L}_\text{total}$ could be the loss in Eq.~\ref{eq:cvae_loss} for CVAE-based model or Eq.~\ref{eq:cgan_loss} for cGAN-based model.

\textbf{Protect benign performance using data augmentation.}
Adversarial training often leads to performance degradation on clean data~\cite{defense:zhang2019theoretically}. However, trajectory prediction is a critical component for safety-critical AD systems and its performance degradation can result in severe consequences (e.g., collisions). Thus, it is important to balance the model performance on the clean and adversarial data when designing adversarial training algorithms.

To further improve the performance on clean and adversarial data,
we need to address the overfitting problem of the min-max adversarial training~\cite{rice2020overfitting}. 
Data augmentation is shown to be effective in addressing the problem in the image classification domain~\cite{rebuffi2021fixing}. 
However, data augmentation in trajectory prediction is rarely studied and non-trivial. To design an effective augmentation algorithm, \citet{rebuffi2021fixing} argues that the most important criterion is that the augmented data should be realistic and diverse. Thus, we design a dynamic-model based data augmentation strategy $\mathbb{A}$ shown in the Appendix A. By using the augmentation, we can generate diverse, realistic multi-agent trajectories for each scene and construct $\mathbb{D}_\text{aug}$. 

\textbf{\method.}
In summary, our adversarial training strategy for trajectory prediction models is formulated as follows:
\begin{equation}\label{eq:robust_traj}
\vspace{-0.2em}
\begin{split}
     \delta =& \argmax_{\delta \in \mathbb{S}} \mathcal{L}_\text{adv}(\,\mathbf{X} + \delta,  \mathbf{Y}), \quad\text{where}  \{\mathbf{X},\mathbf{Y}\} \in \mathbb{D} \cup \mathbb{D}_\text{aug}\\
     \theta,\,\phi  =& \argmin_{ \theta,\,\phi} \mathcal{L}_\text{total}(\mathbf{X}+\delta,\mathbf{Y}) + \mathcal{L}_\text{clean}(\mathbf{X},\mathbf{Y}) + \beta\cdot \mathcal{L}_\text{reg}(\, \mathbf{X},\, \mathbf{X}+\delta\,)\text{, }
\end{split}
\end{equation}
where $\mathbb{D}$, $\mathbb{D}_\text{aug}$ are the training data and augmented data; $\mathcal{L}_\text{adv}$ is  adversarial loss  to generate effective adversarial examples in Eq.~\ref{eq:adv_loss}; $\mathcal{L}_\text{total}$ is the loss in Eq.~\ref{eq:cvae_loss} or Eq.~\ref{eq:cgan_loss} to train a robust model against adversarial examples; $\mathcal{L}_\text{reg}$ and $\mathcal{L}_\text{clean}$ are loss shown in Eq.~\ref{eq:reg}  and Eq.~\ref{eq:clean} to provide a stable signal for training.  $\beta$ is a hyper-parameter for adjusting the regularization.
\section{Experiments and Results}
\subsection{Experimental setup}
\noindent\textbf{Dataset and models. }
We follow the setting in prior work~\cite{yuan2021agent,salzmann2020trajectron++} and use the nuScenes dataset~\cite{caesar2020nuscenes} for evaluation.
For the trajectory prediction models, we select the representative conditional generative models based on CVAE (AgentFormer~\cite{yuan2021agent}) and cGAN (Social-GAN~\cite{alahi2016social}).
AgentFormer is a state-of-the-art model based on CVAE and Social-GAN is a classic model based on cGAN. 
We report the final results for all three models: AgentFormer (AF), mini-AgentFormer (mini-AF) and Social-GAN. More details are shown in the Appendix B.

\noindent\textbf{Training details and hyperparameter choices. }
For the adversarial training, we choose a 2-step Projected Gradient Descent (PGD) attack for the inner maximization and choose $\beta=0.1$. 
We train 50 epochs and 100 epochs for AgentFormer and Social-GAN respectively. For other hyperparameters during training, we follow the original settings for AgentFormer and Social-GAN. 
The details for choosing these hyperparameters can be found in the Appendix B. All experiments are done on the NVIDIA V100 GPU~\cite{NVA100}. 
We consider various baselines, including na\"ive adversarial training (na\"ive AT) and four defenses proposed by Zhang \emph{et al.}~\cite{zhang2022adversarial}: data augmentation with adversarial examples (\textit{DA}), \textit{train-time smoothing}, \textit{test-time smoothing}, \textit{DA + train-time smoothing} and \textit{detection + test-time smoothing}.

\noindent\textbf{Attack and evaluation metrics. }
For the adversarial attack, we choose a 20-step PGD attack (an ablation study on step convergence can be found in the Appendix B).
Without loss of generality, we use $L_\infty$ as the attack threat model so that $\mathbb{S} = \{\delta | \parallel\delta\parallel_\infty \leq \epsilon\}$.
We select $\epsilon=\{ 0.5, 1.0\}$-meter, where the 1-meter deviation is the maximum change for a standard car without shifting to another lanes~\cite{zhang2022adversarial}.
We use four standard evaluation metrics for the nuscenes prediction challenge~\cite{caesar2020nuscenes}: average displacement error (ADE), final displacement error (FDE), off road rates (ORR), and miss rate (MR).
 We evaluate the model's performance on both clean and adversarial data. For convenience, we use \textit{ADE}, \textit{FDE}, \textit{ORR}, \textit{MR} to represent the performance on the clean data and \textit{Robust ADE}, \textit{Robust FDE}, \textit{Robust ORR}, \textit{Robust MR} to represent the performance under attacks. We compute these metrics with the best of five predicted trajectory samples, i.e., $K=5$. 

\subsection{Main results }
Here, we present our main results of \method.
We compare it with the baselines including model trained with clean data (\textit{Clean}) and na\"ive adversarial training (\textit{Na\"ive AT}), and existing defense methods for trajectory prediction~\cite{zhang2022adversarial}. 
The results have been shown in Table~\ref{tab:robust_traj_eval}. 

\begin{table}[t]
\small
\caption{ ADE and Robust ADE on different defense methods and models. The \textcolor{teal}{1-st} and \textcolor{purple}{2-nd} lowest errors are colored.}
\label{tab:robust_traj_eval}
\centering
\adjustbox{max width=\textwidth}{
\begin{tabular}{l|cc|cc|cc|cc|cc|cc}
\toprule
Model         & \multicolumn{4}{c|}{mini-AF} & \multicolumn{4}{c|}{AF} & \multicolumn{4}{c}{SGAN}  \\\midrule
\multirow{2}{*}{Method}         & \multicolumn{2}{c|}{ADE} & \multicolumn{2}{c|}{Robust ADE} & \multicolumn{2}{c|}{ADE} & \multicolumn{2}{c|}{Robust ADE} & \multicolumn{2}{c|}{ADE} & \multicolumn{2}{c}{Robust ADE} \\
                & 0.5        & 1.0        & 0.5            & 1.0           & 0.5        & 1.0        & 0.5            & 1.0      & 0.5        & 1.0        & 0.5            & 1.0    \\\midrule
 Clean          & \color{teal}\textbf{2.05}       & \color{teal}\textbf{2.05}       & 6.86           & 11.53         & \color{teal}\textbf{1.86 }      & \color{teal}\textbf{1.86}       & 5.09           & 8.57  & \color{teal}\textbf{4.80}       & \color{teal}\textbf{4.80}       & 10.52          & 20.15   \\\midrule
                                  \textit{Na\"ive AT~\cite{madry2017towards} }      & 2.75       & 2.78       & 5.44           & 9.20          & 2.52       & 2.56       & \color{purple}\textbf{3.81}           & 6.81             & 6.43       & 6.55       & \color{purple}\textbf{8.34}           & \color{purple}\textbf{14.63} \\
                                  \textit{DA~\cite{zhang2022adversarial} }      & 2.31 & 2.32 & 5.54 & 9.32 & 2.10 & 2.08 & 4.35 & 7.22 & 5.41 & 5.40 & 8.85 & 17.25           \\
                                 \textit{Train-time Smoothing~\cite{zhang2022adversarial} }     & 3.14 & 3.07 & 5.67 & 9.31 & 2.11 & 2.13 & 4.19 & 6.79    & 5.50 & 5.47 & 8.74 & 16.51     \\
                                \textit{Test-time Smoothing~\cite{zhang2022adversarial}}      & 2.97 & 3.07 & \color{purple}\textbf{4.96} & \color{purple}\textbf{8.50} & 2.40 & 2.41 & 4.43 & 7.44      & 6.16 & 6.17 & 9.05 & 17.42        \\
                                  \textit{DA + Train-time Smoothing~\cite{zhang2022adversarial}}      & 2.41       & 2.39       & 5.48           & 9.00          & 2.17       & 2.13       & 4.14           & \color{purple}\textbf{6.62}    & 5.63       & 5.61       & 8.60           & 16.14               \\
                                  \textit{Detection + Test Smoothing~\cite{zhang2022adversarial}} & 2.31       & 2.28       & 5.91           & 9.85          & 2.08       & 2.03       & 4.45           & 7.59  & 5.35       & 5.37       & 9.28           & 17.39          \\
                                  \textbf{\method{} }       & \color{purple}\textbf{2.14}       & \color{purple}\textbf{2.11}       & \color{teal}\textbf{3.69}           & \color{teal}\textbf{3.82}          & \color{purple}\textbf{1.91}       & \color{purple}\textbf{1.95}       & \color{teal}\textbf{2.73}           & \color{teal}\textbf{2.86}      & \color{purple}\textbf{4.95}       & \color{purple}\textbf{5.07}      & \color{teal}\textbf{5.20}           & \color{teal}\textbf{6.94}          \\\bottomrule
\end{tabular}
}
\end{table}

We observe that our method achieves the best robustness and maintains good clean performance for most cases. For instance, with $\epsilon=0.5$ attack on AgentFormer model, our method is able to reduce 46\% prediction errors ($\frac{5.09-2.73}{5.09}$) under the attack at a cost of 2.6\% ($\frac{1.91-1.86}{1.86}$) clean performance degradation on ADE, compared to the model trained with clean data at $\epsilon=0.5$. Compared to the existing methods, our method also significantly outperforms in terms of the robustness. For instance, with $\epsilon=1.0$ attack on AgentFormer model, our method achieves 45\% better robustness with 9\% better clean performance on ADE compared to the best results from existing methods~\cite{zhang2022adversarial}.



\textbf{Impacts to downstream planners.}
To further study the downstream impact of our robust trajectory model in the AD stack, we plug it into a planner. We select a MPC-based planner and evaluate the collision rates under the attack. To perform the attack on a closed-loop planner, we conduct attacks on a sequence of frames with the expectation over transformation (EOT)~\cite{eykholt2018robust} method. We follow the setting from ~\citet{zhang2022adversarial} and choose $\epsilon=1$. We choose AgentFormer model since it has the most competitive performance. As a result, we observe that, while AgentFormer model trained on clean data leads to 10 collision cases under attack, the robust trained model with the proposed \method{} is able to avoid all the collisions. As shown in Fig.~\ref{fig:adv_demo}, we demonstrate that the proposed \method{} is able to avoid the collisions (Fig.~\ref{fig:adv_demo} (d)) while the \textit{DA + Train-time Smoothing} method proposed by \citet{zhang2022adversarial} is not (Fig.~\ref{fig:adv_demo} (c)).

\begin{figure}[t]
    \centering
    \begin{minipage}{.24\textwidth}
        \centering
        \includegraphics[width=\textwidth,trim={25cm 12cm 25cm 10cm},clip]{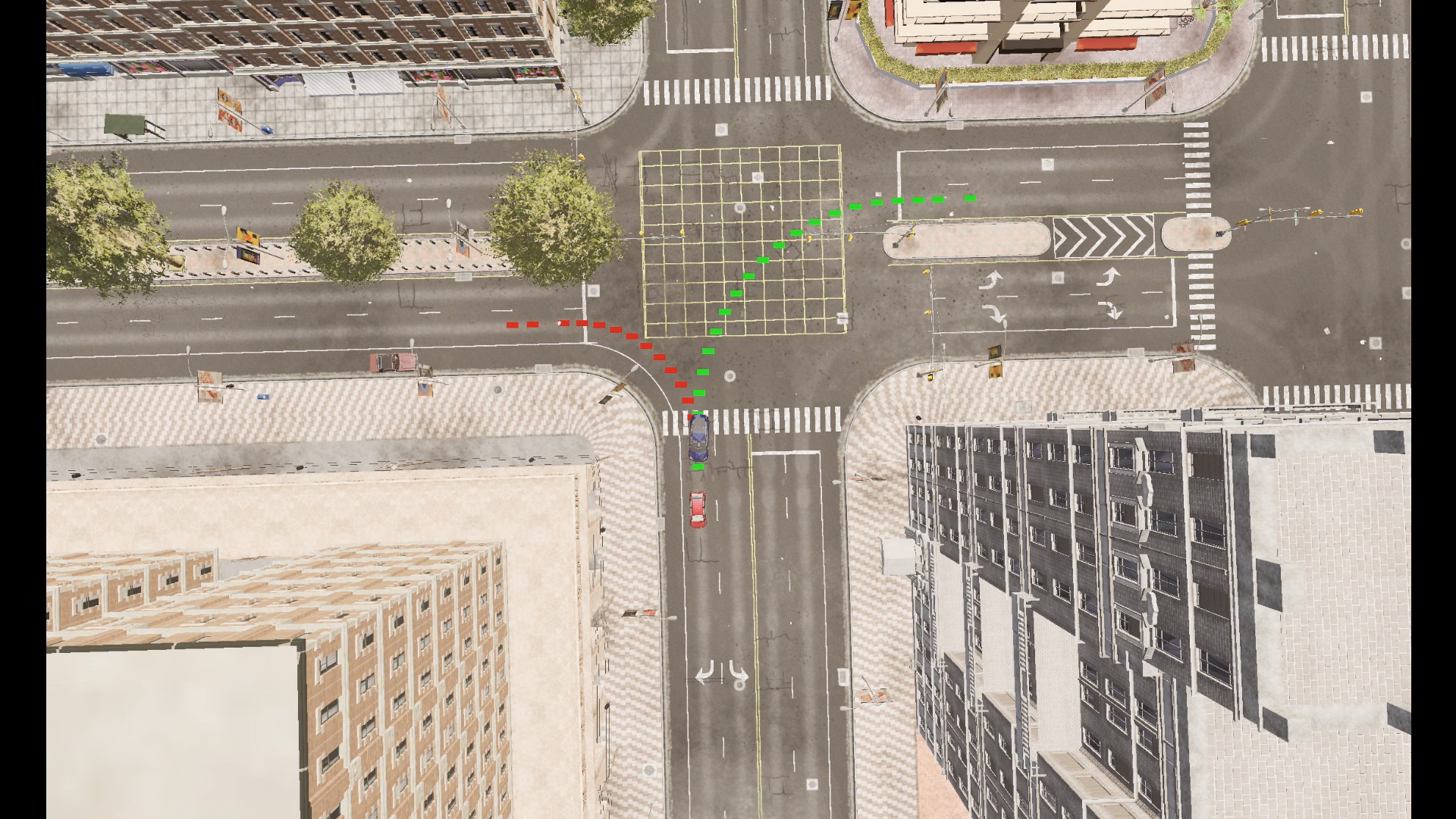}
        \text{\footnotesize(a) Benign case}
    \end{minipage}
    \hfill
    \begin{minipage}{.24\textwidth}
        \centering
        \includegraphics[width=\textwidth,trim={25cm 12cm 25cm 10cm},clip]{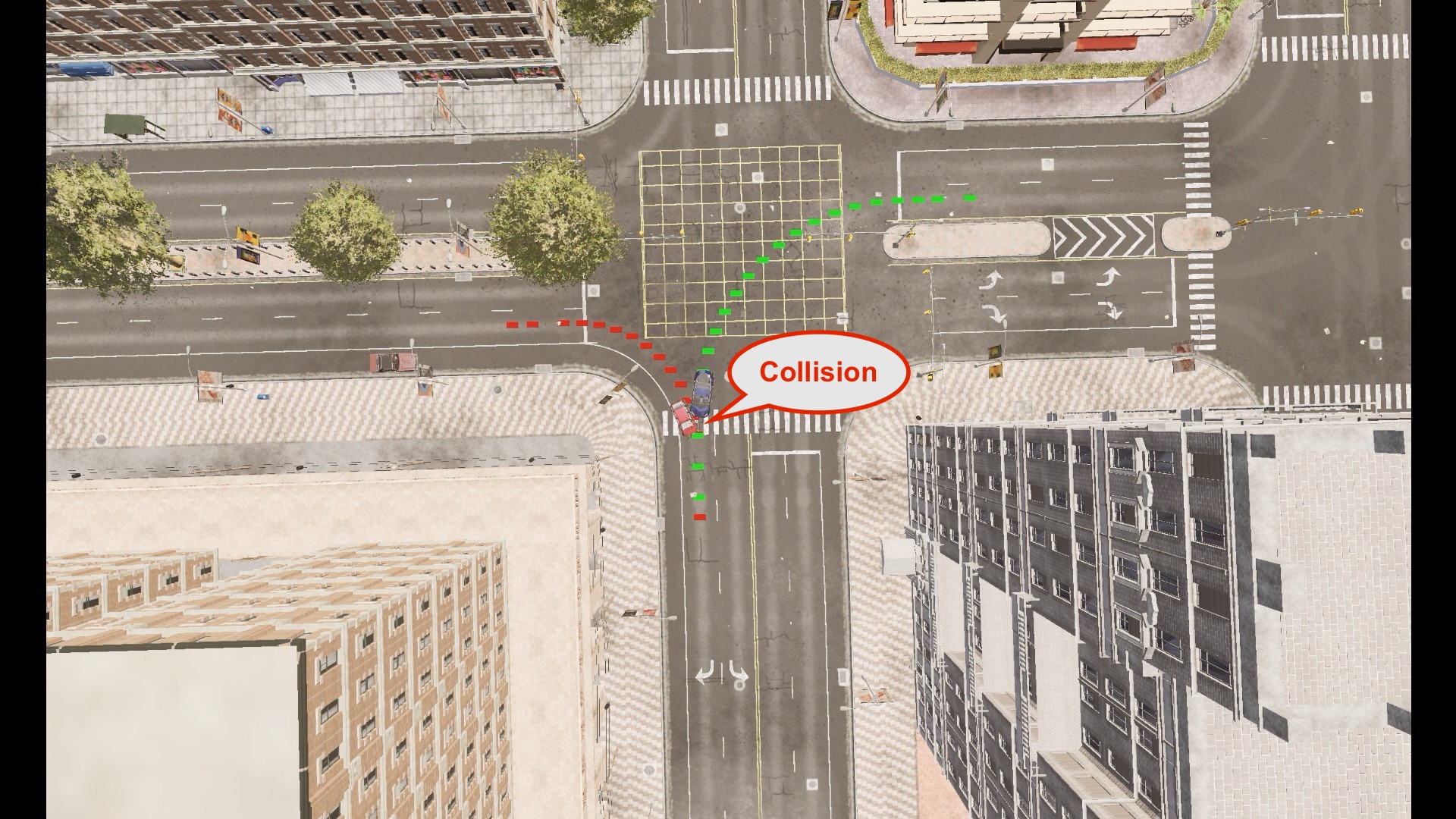}
        \text{\footnotesize(b) Adv Attack}
    \end{minipage}
    \hfill
    \begin{minipage}{.24\textwidth}
        \centering
        \includegraphics[width=\textwidth,trim={25cm 12cm 25cm 10cm},clip]{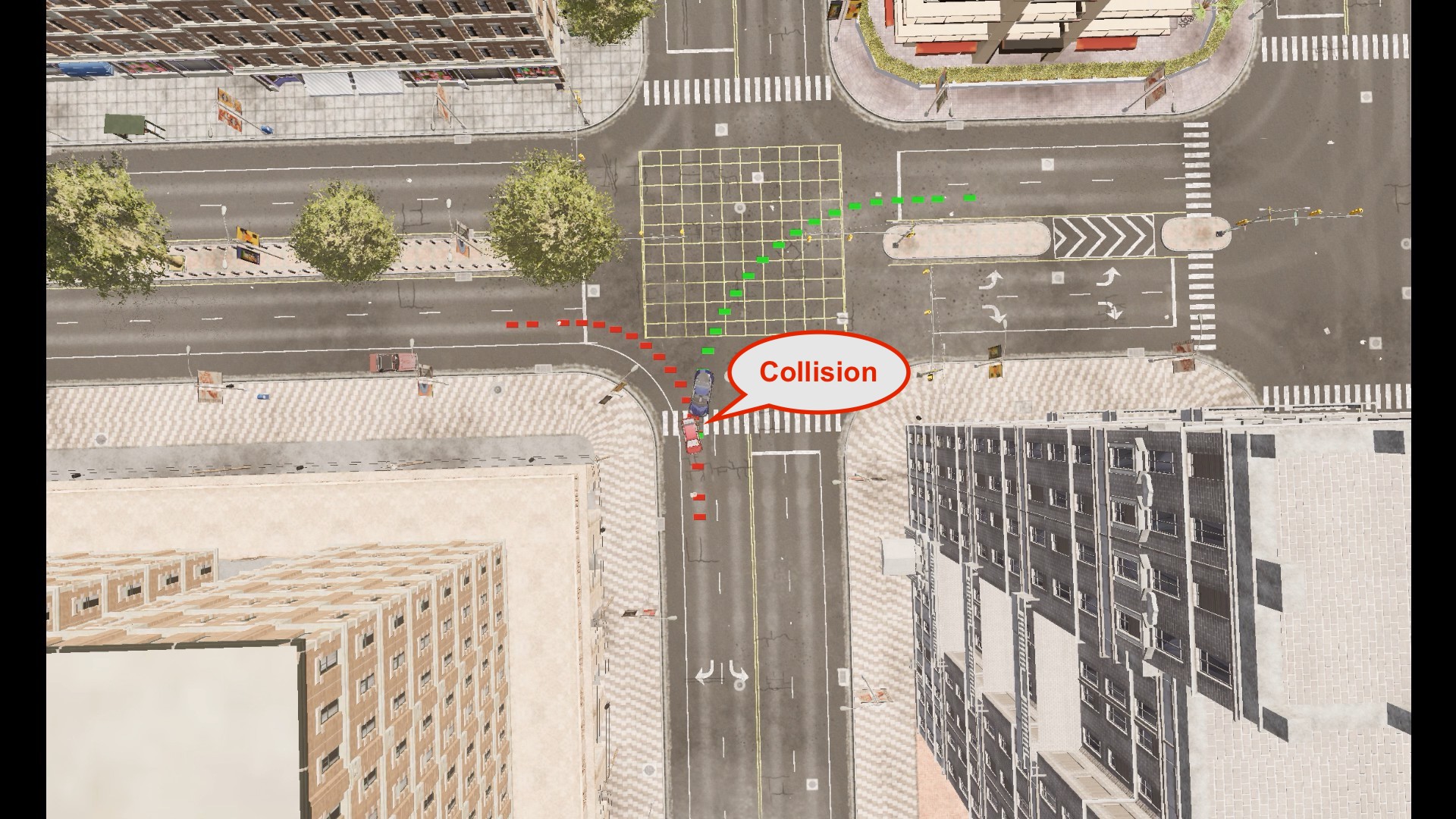}
        \text{\footnotesize(c) w/ defense~\cite{zhang2022adversarial}}
    \end{minipage}
    \hfill
    \begin{minipage}{.24\textwidth}
        \centering
        \includegraphics[width=\textwidth,trim={25cm 12cm 25cm 10cm},clip]{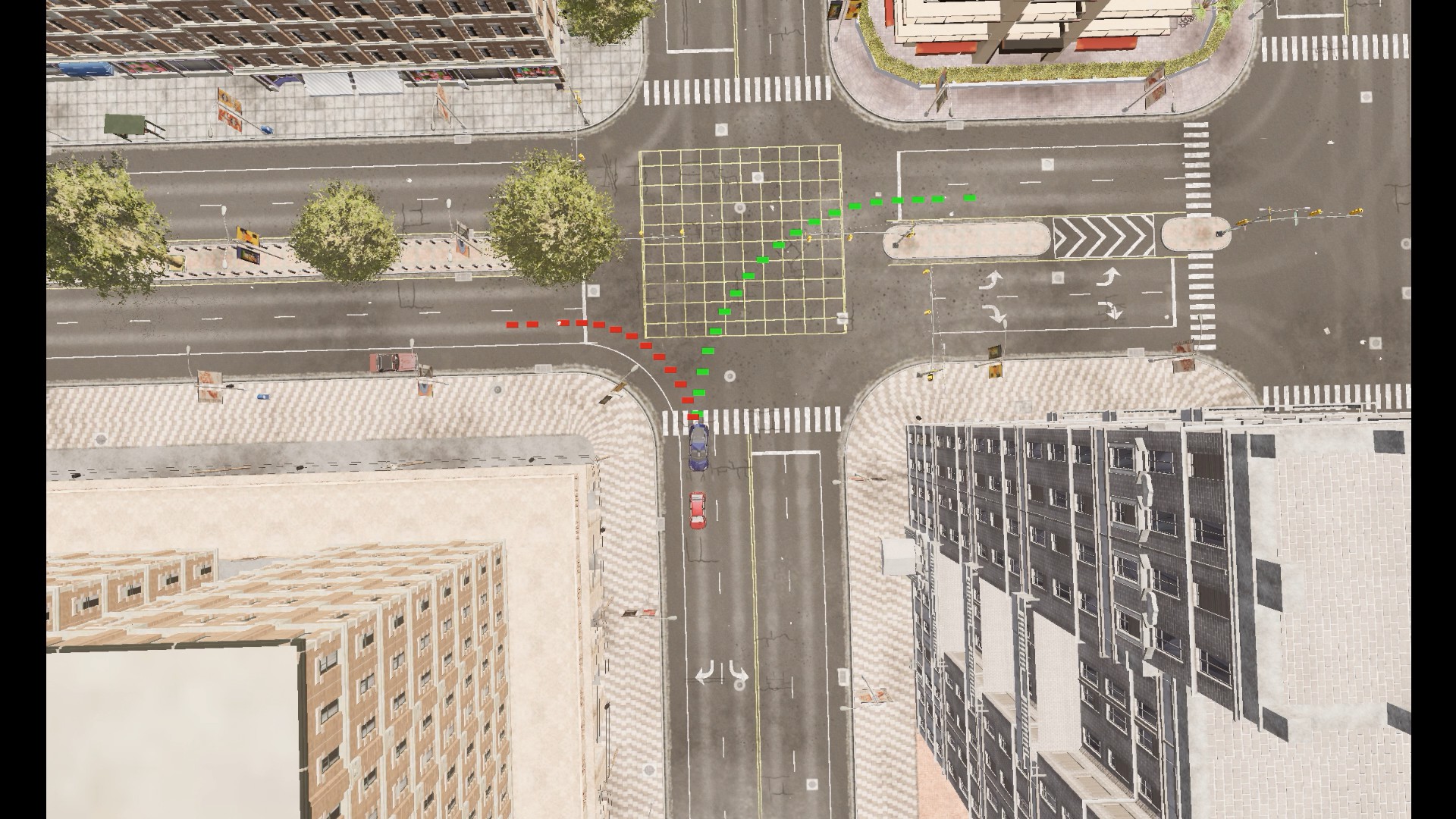}
        \text{\footnotesize(d) w/ defense (ours)}
    \end{minipage}
    \small
        \caption{Impacts to a MPC-basd downstream planner. (a) is under the benign case while (b), (c) and (d) are under the adversarial attacks. 
        The blue car and the red car represent the AV and the adversarial agent respectively.}
        
    \label{fig:adv_demo}
\end{figure}


\subsection{Component analysis}
In this section, we analyze the effectiveness of the three components. 
We use the mini-AgentFormer model since it has competitive performance and is lightweight for a fast adversarial training process.

\textbf{Effectiveness of the \attack.}
To demonstrate the importance of the \attack, we compare it with competitive alternatives, \latent and \context, which also construct the deterministic path.  However, they only attack a partial model as opposed to our end-to-end  full model attack. More details about these attacks are in the Appendix A.
We evaluate their attack effectiveness by attacking a normally trained trajectory prediction model (without robust training).
\begin{wrapfigure}{r}{0.45\textwidth}
    \centering
    \footnotesize
    \vspace{-1.5em}
    \includegraphics[width=0.42\textwidth]{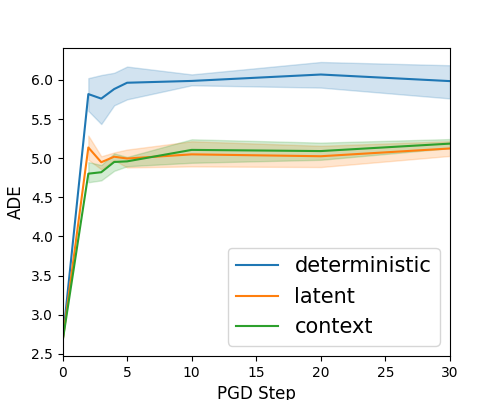}
    \caption{Peformance of different attacks in mini-AgenetFormer.}
    \vspace{-1.5em}
    \label{fig:analysis_attack_loss}
\end{wrapfigure}
In Fig.~\ref{fig:analysis_attack_loss}, we demonstrate that \attack is the most effective attack among all. 
Additionally, 
we embed them into the whole adversarial training pipeline and evaluate the adversarial robustness.
The results are shown in Table~\ref{tab:adv_analysis}. We observe that the model trained with the \attack{} achieves the best robustness in terms of ADE.  More results with other metrics and the other $\epsilon$ are in the Appendix B.

\begin{wraptable}{r}{0.6\textwidth}
        \vspace{-1em}
        \small
        \caption{
        ADE and robust ADE for different methods on mini-AgentFormer. The lowest error is in bold. 
        }
        \label{tab:adv_analysis}
        \centering
        \begin{tabular}{l|cc|cc}
            \toprule
            Method                                     & \multicolumn{2}{c|}{ADE} & \multicolumn{2}{c}{Robust ADE} \\
                                                      & 0.5 & 1.0 & 0.5       & 1.0  \\\midrule 
            Clean                          & 2.05 & 2.05 & 6.86       & 11.53 \\\midrule
            \latent{}                                    & 2.55 & 2.70 & 4.10       & 4.71  \\
            \context{}                                   & \textbf{2.47} & 2.59 & 3.94       & 4.78 \\
            \attack{}                                    & 2.61 & \textbf{2.55} & \textbf{3.88}       & \textbf{4.35}  \\\midrule
            \attack{}                                    &  &  &        &  \\
            + $\mathcal{L}_\text{reg}$                              & 2.29	&2.31	&	3.76 &	4.28	\\
             + $\mathcal{L}_{\text{clean}}$ + $\mathcal{L}_\text{reg}$                        & 2.23 & 2.19 & 3.71       & 3.83\\
             + $\mathcal{L}_{\text{clean}}$ + $\mathcal{L}_\text{reg}$ + Aug                  & \textbf{2.14} & {\bf 2.11} & {\bf 3.69}       & {\bf 3.82} \\
             \bottomrule
        \end{tabular}
        \vspace{-3em}
\end{wraptable}  
\textbf{Effect of additional loss functions.}
We evaluate the performance of the models trained with additional loss terms: $\mathcal{L}_{\text{clean}}$ and $\mathcal{L}_{\text{reg}}$. In Table~\ref{tab:adv_analysis}, 
we can see that the regularization term $\mathcal{L}_{\text{reg}}$ improves robustness of the models and achieves better clean performance. It shows that the regularization of the introduced noises on conditional variables help the model to stabilize the training procedure. By adding the clean loss $\mathcal{L}_\text{clean}$, we observe that both the robustness and clean performance are improved further, which means the benign data indeed anchors the model output on clean data distribution and provides a stable signal for the better robust training for generative models.

\textbf{Effect of domain-specific augmentation.}
To demonstrate the effectiveness of the domain-specific augmentation, 
We also combine it with all of the above components to validate its effect. The results are shown in Table~\ref{tab:adv_analysis}. We observe that it achieves a better performance on clean and adversarial data. 

\section{Limitations}
In this work, we identified the challenges of applying adversarial training on trajectory prediction models based on conditional generative models. Though the conditional generative model is the main-stream architecture for the trajectory prediction task, there are other architectures (e.g., flow-based method~\cite{rhinehart2018r2p2,rhinehart2019precog} and RL-based method~\cite{deo2020trajectory}) for generating multi-modal predictions. 
We leave these as future works for building robust trajectory prediction models.
\section{Conclusion}
In this paper, we aim to study how to train robust generative trajectory prediction models against adversarial attacks, which is seldom explored in the literature. 
To achieve this goal, we first identify three key challenges in designing an adversarial training framework to train robust trajectory prediction models. To address them, we propose an adversarial training framework with three main components, including (1) a \textit{deterministic attack} for the inner maximization process of the adversarial training, (2) additional regularization terms for stable outer minimization of adversarial training, and (3) a domain-specific augmentation strategy to achieve a better performance trade-off on clean and adversarial data.
To show the generality of our method, we apply our approach to two trajectory prediction models, including (1) a CVAE-based model, AgentFormer, and (2) a cGAN-based model, Social-GAN. Our extensive experiments show our method could significantly improve the robustness with a slight performance degradation on the clean data, compared to the existing techniques and dramatically reduce the severe collision rates when plugged into the AD stack with a planner. We hope our work can shed light on developing robust trajectory prediction systems for AD.      


{\small
\bibliography{example}

\begin{thebibliography}{36}
\providecommand{\natexlab}[1]{#1}
\providecommand{\url}[1]{\texttt{#1}}
\expandafter\ifx\csname urlstyle\endcsname\relax
  \providecommand{\doi}[1]{doi: #1}\else
  \providecommand{\doi}{doi: \begingroup \urlstyle{rm}\Url}\fi

\bibitem[Alahi et~al.(2016)Alahi, Goel, Ramanathan, Robicquet, Fei-Fei, and
  Savarese]{alahi2016social}
A.~Alahi, K.~Goel, V.~Ramanathan, A.~Robicquet, L.~Fei-Fei, and S.~Savarese.
\newblock Social lstm: Human trajectory prediction in crowded spaces.
\newblock In \emph{Proceedings of the IEEE conference on computer vision and
  pattern recognition}, pages 961--971, 2016.

\bibitem[Ivanovic and Pavone(2019)]{ivanovic2019trajectron}
B.~Ivanovic and M.~Pavone.
\newblock The trajectron: Probabilistic multi-agent trajectory modeling with
  dynamic spatiotemporal graphs.
\newblock In \emph{Proceedings of the IEEE/CVF International Conference on
  Computer Vision}, pages 2375--2384, 2019.

\bibitem[Salzmann et~al.(2020)Salzmann, Ivanovic, Chakravarty, and
  Pavone]{salzmann2020trajectron++}
T.~Salzmann, B.~Ivanovic, P.~Chakravarty, and M.~Pavone.
\newblock Trajectron++: Dynamically-feasible trajectory forecasting with
  heterogeneous data.
\newblock In \emph{European Conference on Computer Vision}, pages 683--700.
  Springer, 2020.

\bibitem[Yuan et~al.(2021)Yuan, Weng, Ou, and Kitani]{yuan2021agent}
Y.~Yuan, X.~Weng, Y.~Ou, and K.~Kitani.
\newblock Agentformer: Agent-aware transformers for socio-temporal multi-agent
  forecasting.
\newblock In \emph{Proceedings of the IEEE/CVF International Conference on
  Computer Vision (ICCV)}, 2021.

\bibitem[Rhinehart et~al.(2018)Rhinehart, Kitani, and
  Vernaza]{rhinehart2018r2p2}
N.~Rhinehart, K.~M. Kitani, and P.~Vernaza.
\newblock R2p2: A reparameterized pushforward policy for diverse, precise
  generative path forecasting.
\newblock In \emph{Proceedings of the European Conference on Computer Vision
  (ECCV)}, pages 772--788, 2018.

\bibitem[Rhinehart et~al.(2019)Rhinehart, McAllister, Kitani, and
  Levine]{rhinehart2019precog}
N.~Rhinehart, R.~McAllister, K.~Kitani, and S.~Levine.
\newblock Precog: Prediction conditioned on goals in visual multi-agent
  settings.
\newblock In \emph{Proceedings of the IEEE/CVF International Conference on
  Computer Vision}, pages 2821--2830, 2019.

\bibitem[Kosaraju et~al.(2019)Kosaraju, Sadeghian, Mart{\'\i}n-Mart{\'\i}n,
  Reid, Rezatofighi, and Savarese]{kosaraju2019social}
V.~Kosaraju, A.~Sadeghian, R.~Mart{\'\i}n-Mart{\'\i}n, I.~Reid, H.~Rezatofighi,
  and S.~Savarese.
\newblock Social-bigat: Multimodal trajectory forecasting using bicycle-gan and
  graph attention networks.
\newblock \emph{Advances in Neural Information Processing Systems}, 32, 2019.

\bibitem[Madry et~al.(2017)Madry, Makelov, Schmidt, Tsipras, and
  Vladu]{madry2017towards}
A.~Madry, A.~Makelov, L.~Schmidt, D.~Tsipras, and A.~Vladu.
\newblock Towards deep learning models resistant to adversarial attacks.
\newblock \emph{arXiv preprint arXiv:1706.06083}, 2017.

\bibitem[Szegedy et~al.(2013)Szegedy, Zaremba, Sutskever, Bruna, Erhan,
  Goodfellow, and Fergus]{szegedy2013intriguing}
C.~Szegedy, W.~Zaremba, I.~Sutskever, J.~Bruna, D.~Erhan, I.~Goodfellow, and
  R.~Fergus.
\newblock Intriguing properties of neural networks.
\newblock \emph{arXiv preprint arXiv:1312.6199}, 2013.

\bibitem[Carlini and Wagner(2017)]{carlini2017towards}
N.~Carlini and D.~Wagner.
\newblock Towards evaluating the robustness of neural networks.
\newblock In \emph{2017 ieee symposium on security and privacy (sp)}, pages
  39--57. IEEE, 2017.

\bibitem[Zhang et~al.(2022)Zhang, Hu, Sun, Chen, and Mao]{zhang2022adversarial}
Q.~Zhang, S.~Hu, J.~Sun, Q.~A. Chen, and Z.~M. Mao.
\newblock On adversarial robustness of trajectory prediction for autonomous
  vehicles.
\newblock \emph{arXiv preprint arXiv:2201.05057}, 2022.

\bibitem[Jeddi et~al.(2020)Jeddi, Shafiee, and Wong]{defense:jeddi2020simple}
A.~Jeddi, M.~J. Shafiee, and A.~Wong.
\newblock A simple fine-tuning is all you need: Towards robust deep learning
  via adversarial fine-tuning.
\newblock \emph{arXiv preprint arXiv:2012.13628}, 2020.

\bibitem[Liu et~al.(2019)Liu, Yu, and Su]{defense:liu2019extending}
D.~Liu, R.~Yu, and H.~Su.
\newblock Extending adversarial attacks and defenses to deep 3d point cloud
  classifiers.
\newblock In \emph{2019 IEEE International Conference on Image Processing
  (ICIP)}, pages 2279--2283. IEEE, 2019.

\bibitem[Papernot et~al.(2016)Papernot, McDaniel, Wu, Jha, and
  Swami]{defense:papernot2016distillation}
N.~Papernot, P.~McDaniel, X.~Wu, S.~Jha, and A.~Swami.
\newblock Distillation as a defense to adversarial perturbations against deep
  neural networks.
\newblock In \emph{2016 IEEE symposium on security and privacy (SP)}, pages
  582--597. IEEE, 2016.

\bibitem[Papernot and McDaniel(2017)]{defense:papernot2017extending}
N.~Papernot and P.~McDaniel.
\newblock Extending defensive distillation.
\newblock \emph{arXiv preprint arXiv:1705.05264}, 2017.

\bibitem[Shafahi et~al.(2019)Shafahi, Najibi, Ghiasi, Xu, Dickerson, Studer,
  Davis, Taylor, and Goldstein]{defense:shafahi2019adversarial}
A.~Shafahi, M.~Najibi, A.~Ghiasi, Z.~Xu, J.~Dickerson, C.~Studer, L.~S. Davis,
  G.~Taylor, and T.~Goldstein.
\newblock Adversarial training for free!
\newblock \emph{arXiv preprint arXiv:1904.12843}, 2019.

\bibitem[Wong et~al.(2020)Wong, Rice, and Kolter]{defense:wong2020fast}
E.~Wong, L.~Rice, and J.~Z. Kolter.
\newblock Fast is better than free: Revisiting adversarial training.
\newblock \emph{arXiv preprint arXiv:2001.03994}, 2020.

\bibitem[Xie et~al.(2020)Xie, Tan, Gong, Wang, Yuille, and
  Le]{defense:xie2020adversarial}
C.~Xie, M.~Tan, B.~Gong, J.~Wang, A.~L. Yuille, and Q.~V. Le.
\newblock Adversarial examples improve image recognition.
\newblock In \emph{Proceedings of the IEEE/CVF Conference on Computer Vision
  and Pattern Recognition}, pages 819--828, 2020.

\bibitem[Xie and Yuille(2020)]{defense:Xie2020Intriguing}
C.~Xie and A.~Yuille.
\newblock Intriguing properties of adversarial training at scale.
\newblock In \emph{International Conference on Learning Representations}, 2020.
\newblock URL \url{https://openreview.net/forum?id=HyxJhCEFDS}.

\bibitem[Xie et~al.(2020)Xie, Tan, Gong, Yuille, and Le]{defense:xie2020smooth}
C.~Xie, M.~Tan, B.~Gong, A.~Yuille, and Q.~V. Le.
\newblock Smooth adversarial training.
\newblock \emph{arXiv preprint arXiv:2006.14536}, 2020.

\bibitem[Xu et~al.(2017)Xu, Evans, and Qi]{defense:xu2017feature}
W.~Xu, D.~Evans, and Y.~Qi.
\newblock Feature squeezing: Detecting adversarial examples in deep neural
  networks.
\newblock \emph{arXiv preprint arXiv:1704.01155}, 2017.

\bibitem[Yang et~al.(2019)Yang, Zhang, Katabi, and Xu]{defense:yang2019me}
Y.~Yang, G.~Zhang, D.~Katabi, and Z.~Xu.
\newblock Me-net: Towards effective adversarial robustness with matrix
  estimation.
\newblock \emph{arXiv preprint arXiv:1905.11971}, 2019.

\bibitem[Zhang et~al.(2019)Zhang, Yu, Jiao, Xing, El~Ghaoui, and
  Jordan]{defense:zhang2019theoretically}
H.~Zhang, Y.~Yu, J.~Jiao, E.~Xing, L.~El~Ghaoui, and M.~Jordan.
\newblock Theoretically principled trade-off between robustness and accuracy.
\newblock In \emph{International Conference on Machine Learning}, pages
  7472--7482. PMLR, 2019.

\bibitem[Athalye et~al.(2018)Athalye, Carlini, and
  Wagner]{athalye2018obfuscated}
A.~Athalye, N.~Carlini, and D.~Wagner.
\newblock Obfuscated gradients give a false sense of security: Circumventing
  defenses to adversarial examples.
\newblock In \emph{International Conference on Machine Learning}, pages
  274--283. PMLR, 2018.

\bibitem[Deng(2012)]{deng2012mnist}
L.~Deng.
\newblock The mnist database of handwritten digit images for machine learning
  research.
\newblock \emph{IEEE Signal Processing Magazine}, 29\penalty0 (6):\penalty0
  141--142, 2012.

\bibitem[{Wikipedia contributors}(2022)]{saltpepper-wiki:1085283109}
{Wikipedia contributors}.
\newblock Salt-and-pepper noise --- {Wikipedia}{,} the free encyclopedia, 2022.
\newblock URL
  \url{https://en.wikipedia.org/w/index.php?title=Salt-and-pepper_noise&oldid=1085283109}.
\newblock [Online; accessed 16-June-2022].

\bibitem[Rice et~al.(2020)Rice, Wong, and Kolter]{rice2020overfitting}
L.~Rice, E.~Wong, and Z.~Kolter.
\newblock Overfitting in adversarially robust deep learning.
\newblock In \emph{International Conference on Machine Learning}, pages
  8093--8104. PMLR, 2020.

\bibitem[Rebuffi et~al.(2021)Rebuffi, Gowal, Calian, Stimberg, Wiles, and
  Mann]{rebuffi2021fixing}
S.-A. Rebuffi, S.~Gowal, D.~A. Calian, F.~Stimberg, O.~Wiles, and T.~Mann.
\newblock Fixing data augmentation to improve adversarial robustness.
\newblock \emph{arXiv preprint arXiv:2103.01946}, 2021.

\bibitem[Caesar et~al.(2020)Caesar, Bankiti, Lang, Vora, Liong, Xu, Krishnan,
  Pan, Baldan, and Beijbom]{caesar2020nuscenes}
H.~Caesar, V.~Bankiti, A.~H. Lang, S.~Vora, V.~E. Liong, Q.~Xu, A.~Krishnan,
  Y.~Pan, G.~Baldan, and O.~Beijbom.
\newblock nuscenes: A multimodal dataset for autonomous driving.
\newblock In \emph{Proceedings of the IEEE/CVF conference on computer vision
  and pattern recognition}, pages 11621--11631, 2020.

\bibitem[Choquette et~al.(2021)Choquette, Gandhi, Giroux, Stam, and
  Krashinsky]{NVA100}
J.~Choquette, W.~Gandhi, O.~Giroux, N.~Stam, and R.~Krashinsky.
\newblock Nvidia a100 tensor core gpu: Performance and innovation.
\newblock \emph{IEEE Micro}, 41\penalty0 (2):\penalty0 29--35, 2021.
\newblock \doi{10.1109/MM.2021.3061394}.

\bibitem[Eykholt et~al.(2018)Eykholt, Evtimov, Fernandes, Li, Rahmati, Xiao,
  Prakash, Kohno, and Song]{eykholt2018robust}
K.~Eykholt, I.~Evtimov, E.~Fernandes, B.~Li, A.~Rahmati, C.~Xiao, A.~Prakash,
  T.~Kohno, and D.~Song.
\newblock Robust physical-world attacks on deep learning visual classification.
\newblock In \emph{Proceedings of the IEEE Conference on Computer Vision and
  Pattern Recognition}, pages 1625--1634, 2018.

\bibitem[Deo and Trivedi(2020)]{deo2020trajectory}
N.~Deo and M.~M. Trivedi.
\newblock Trajectory forecasts in unknown environments conditioned on
  grid-based plans.
\newblock \emph{arXiv preprint arXiv:2001.00735}, 2020.

\bibitem[Polack et~al.(2017)Polack, Altché, d'Andréa Novel, and
  de~La~Fortelle]{MATUTE2019kinematic}
P.~Polack, F.~Altché, B.~d'Andréa Novel, and A.~de~La~Fortelle.
\newblock The kinematic bicycle model: A consistent model for planning feasible
  trajectories for autonomous vehicles?
\newblock In \emph{2017 IEEE Intelligent Vehicles Symposium (IV)}, pages
  812--818, 2017.
\newblock \doi{10.1109/IVS.2017.7995816}.

\bibitem[McNaughton et~al.(2011)McNaughton, Urmson, Dolan, and
  Lee]{mcnaughton2011motion}
M.~McNaughton, C.~Urmson, J.~M. Dolan, and J.-W. Lee.
\newblock Motion planning for autonomous driving with a conformal
  spatiotemporal lattice.
\newblock In \emph{2011 IEEE International Conference on Robotics and
  Automation}, pages 4889--4895. IEEE, 2011.

\bibitem[Camacho and Alba(2013)]{camacho2013model}
E.~F. Camacho and C.~B. Alba.
\newblock \emph{Model predictive control}.
\newblock Springer science \& business media, 2013.

\bibitem[Croce and Hein(2020)]{croce2020reliable}
F.~Croce and M.~Hein.
\newblock Reliable evaluation of adversarial robustness with an ensemble of
  diverse parameter-free attacks.
\newblock In \emph{International Conference on Machine Learning}, pages
  2206--2216. PMLR, 2020.

\end{thebibliography}
}
\appendix
\section{Method and Implementations}
\subsection{Adversarial Attack on Trajectory Prediction}
\textbf{\latent and \context. }
Noticed that, besides \attack introduced in the main paper, there are also two other less intuitive attacks. Since the prediction $\hat{\mathbf{Y}}$ is dependent on posterior distribution $q_\phi(\mathbf{Z}|\mathbf{X},\mathbf{Y})$ and conditional variable $f(\mathbf{X})$, we can construct attacks based on that. \textit{Latent attack} aims to increase the error of estimating $q_\phi(\mathbf{Z}|\mathbf{X},\mathbf{Y})$, which is formulated as:
\begin{equation}
    \delta = \argmax_{\delta} \text{KL}(\, q_\phi(\mathbf{Z} | \mathbf{Y},\mathbf{X}) \parallel q_\phi(\mathbf{Z} | \mathbf{Y},\mathbf{X} + \delta) \,)\text{. }
\end{equation}
\textit{Context attack} aims to increase the error of encoding the conditional variable $f(Z)$, which is formulated as:
\begin{equation}
    \delta = \argmax_{\delta} \mathrm{d}(\, f(\mathbf{X}),\, f(\mathbf{X}+\delta) \,) \text{, }
\end{equation}
where $\mathrm{d}$ is a distance function (e.g., $L_2$ norm).
However, \textit{latent attack} and \textit{context attack} are effective due to two reasons. First, they are exploiting the vulnerability of a partial model. For example, \textit{latent attack} only exploits the posterior estimation $q_\phi(\mathbf{Z}|\mathbf{X},\mathbf{Y})$ and \textit{context attack} only exploits the conditional encoder $f(\mathbf{X})$. Second, these attacks aim for a different goal. For the \textit{latent attack} and \textit{context attack}, the objectives are set for finding adversarial perturbations that maximize the difference of generated posterior distribution/context given $\mathbf{X}$ and $\mathbf{X}+\delta$, due to lacking ground truth for intermediate latent variables. However, for the sample attack, the objective is directly set for maximizing the prediction errors from the ground truth (future trajectories), which is more effective.

\textbf{Adversarial attack on consecutive frames. }
In order to fool a planner in a closed-loop manner to make consistent wrong decisions, we need to conduct adversarial attacks on consecutive frames. To attack $L_p$ consecutive frames of predictions, we aim to generate the adversarial trajectory of length $H+L_p$ that uniformly misleads the prediction at each time frame. 
To achieve this goal, we can easily extend the formulation for attacking single-step predictions to attack a sequence of predictions, which is useful for attacking a sequence of decision made by AV planning module.
Concretely, to generate the adversarial trajectories for $L_p$ consecutive steps of predictions, we aggregate the adversarial losses over these frames. The objective for attacking a length of $H+L_p$ trajectory is:
\begin{equation}
\label{eq:seq_attack}
    \sum_{t \in [-L_p,\dotsc 0]} \mathcal{L}_{\text{adv}}(\mathbf{X}_\text{adv}(t),\mathbf{Y}(t)) \text{, }
\end{equation}
where $\mathbf{X}_\text{adv}(t),\mathbf{Y}(t)$ are the corresponding $\mathbf{X}+\delta,\mathbf{Y}$ at time frame t.
\subsection{Adversarial Training on Generative Models}
\paragraph{Challenges. }
One challenge that hinders the adversarial training process is the noisy conditional data distribution disturbing the training process. One hypothesis we mentioned in the main paper is, the context encoding can magnify the bounded perturbation $\delta$ on history trajectory $\mathbf{X}$ to an unbounded perturbation on the conditional variable $\mathbf{C} = f(\mathbf{X}+\delta)$, during the training process.
\begin{lemma}\label{lm:lipsch}For a neural network $f$ which is not bounded on Lipschitz constant during the training procedure, given any constant $\eta$ and an input $\mathbf{X}$, there exists a pair $(\delta, f)$, that satisfies $$ \parallel f(\mathbf{X} + \delta) - f(\mathbf{X})\parallel \geq \eta \text{.}$$\end{lemma}
Lemma~\ref{lm:lipsch} can be easily derived by the definition of Lipshitz constant. Lipshitz constant $L$ is defined as
$$ L := \sup \frac{\parallel f(\mathbf{X} + \delta) - f(\mathbf{X})\parallel}{\parallel \delta \parallel} \text{.}$$
If $L$ is not bounded, $\frac{\parallel f(\mathbf{X} + \delta) - f(\mathbf{X})\parallel}{\parallel \delta \parallel}$ is not bounded and so is $\parallel f(\mathbf{X} + \delta) - f(\mathbf{X})\parallel$ given a bounded $\delta$.
This means that, a bounded perturbation $\delta$ can potentially be magnified to be noisier on encoded conditional variable $\mathbf{C} = f(\mathbf{X} + \delta)$.

\begin{figure}[t]
    \centering
    \begin{minipage}{.19\textwidth}
        \centering
        \includegraphics[width=\textwidth]{figs/sample_all.png}
        \text{\footnotesize $p=0$}
    \end{minipage}
    \begin{minipage}{.19\textwidth}
        \centering
        \includegraphics[width=\textwidth]{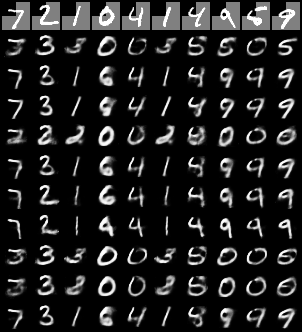}
        \text{\footnotesize $p=0.1$}
    \end{minipage}
    \hfill
    \begin{minipage}{.19\textwidth}
        \centering
        \includegraphics[width=\textwidth]{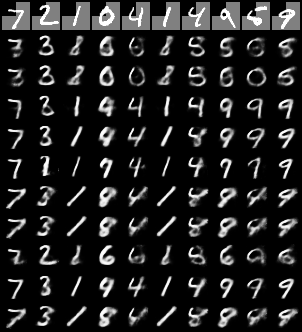}
        \text{\footnotesize $p=0.2$}
    \end{minipage}
    \hfill
    \begin{minipage}{.19\textwidth}
        \centering
        \includegraphics[width=\textwidth]{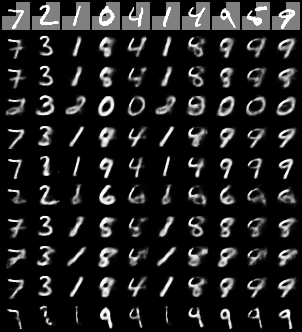}
        \text{\footnotesize $p=0.3$}
    \end{minipage}
    \hfill
    \begin{minipage}{.19\textwidth}
        \centering
        \includegraphics[width=\textwidth]{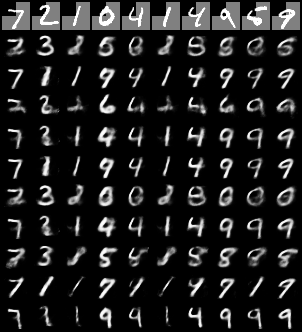}
        \text{\footnotesize $p=0.4$}
    \end{minipage}\\
    \vspace{0.5em}
    \begin{minipage}{.19\textwidth}
        \centering
        \includegraphics[width=\textwidth]{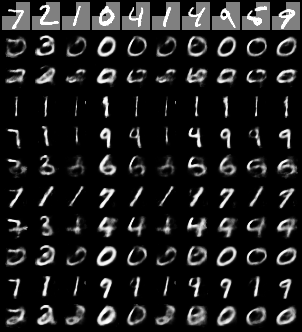}
        \text{\footnotesize $p=0.5$}
    \end{minipage}
    \hfill
    \begin{minipage}{.19\textwidth}
        \centering
        \includegraphics[width=\textwidth]{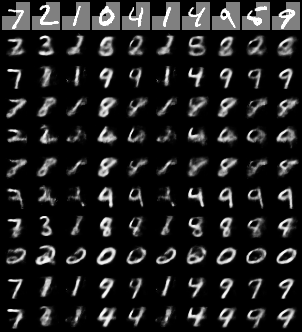}
        \text{\footnotesize $p=0.6$}
    \end{minipage}
    \hfill
    \begin{minipage}{.19\textwidth}
        \centering
        \includegraphics[width=\textwidth]{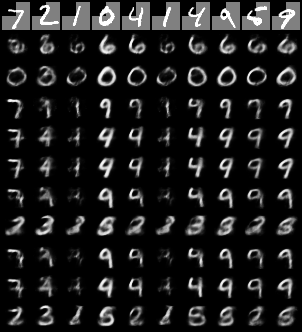}
        \text{\footnotesize $p=0.7$}
    \end{minipage}
    \hfill
    \begin{minipage}{.19\textwidth}
        \centering
        \includegraphics[width=\textwidth]{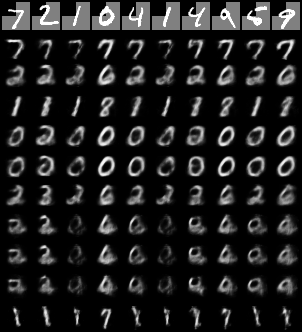}
        \text{\footnotesize $p=0.8$}
    \end{minipage}
    \hfill
    \begin{minipage}{.19\textwidth}
        \centering
        \includegraphics[width=\textwidth]{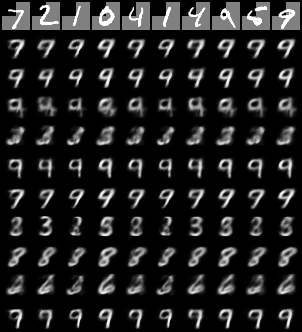}
        \text{\footnotesize $p=0.9$}
    \end{minipage}
        \caption{Visual examples of images generated from models trained with different levels of salt and pepper noises.}
        
    \label{fig:noise_demo}
\end{figure}

\paragraph{Analysis.}
In order to provide a quantitative analysis of the degeneration degree from conditional generative model to generative model (e.g., CVAE to VAE) with respect to the noise level, we propose a method to estimate the correlation between the degeneration and the noise level. Specifically, we trained a classifier with a 2-layer CNN achieving 99\% accuracy on MNIST dataset. 
Then, given a conditional variable (the upper left quarter of an image of digit $y$), we generate images with the conditional generative models and use the classifier to calculate the confidence of the generated images labeled as digit $y$.
We calculate the average confidence of 10 generated images on all 10,000 images in the MNIST test data set. The lower the score means the weaker the correlation between the generated images with the given conditions, or the stronger correlation between the degeneration and the noise level. 
We also provide visualization examples of generated images from models trained using data with different level of noises in Figure~\ref{fig:noise_demo} and Figure~\ref{fig:adv_demo}. We can see that, with the noise level increases, the generated images are less dependent on the conditional variable (i.e., not being the same digit). In the extreme case of high level noise, for example when $p=0.9$, the model generates images solely depends on the random prior value it samples and generates the similar images for each row. We can also see that, the adversarial noises are more effective compared to the salt and pepper noise. With a small amount of noise (i.e., $\epsilon=0.1$), it can degenerate the conditional generative model to a generative model (i.e., CVAE to VAE here).

\begin{figure}
   \begin{minipage}{.19\textwidth}
        \centering
        \includegraphics[width=\textwidth]{figs/sample_all.png}
        \text{\footnotesize $\epsilon=0$}
    \end{minipage}
    \hfill
    \begin{minipage}{.19\textwidth}
        \centering
        \includegraphics[width=\textwidth]{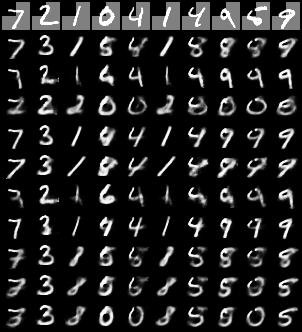}
        \text{\footnotesize $\epsilon=0.01$}
    \end{minipage}
    \hfill
    \begin{minipage}{.19\textwidth}
        \centering
        \includegraphics[width=\textwidth]{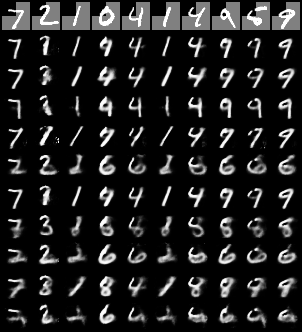}
        \text{\footnotesize $\epsilon=0.03$}
    \end{minipage}
    \hfill
    \begin{minipage}{.19\textwidth}
        \centering
        \includegraphics[width=\textwidth]{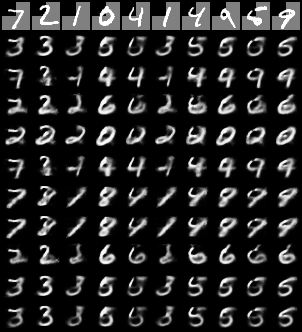}
        \text{\footnotesize $\epsilon=0.05$}
    \end{minipage}
    \hfill
    \begin{minipage}{.19\textwidth}
        \centering
        \includegraphics[width=\textwidth]{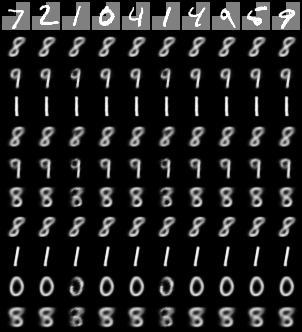}
        \text{\footnotesize $\epsilon=0.1$}
    \end{minipage}
    \hfill
        \caption{Visual examples of images generated from models trained with different levels of adversarial noises.}
        
    \label{fig:adv_demo}
\end{figure}

\subsection{Data Augmentation with Dynamic Model}
For the data augmentation strategy $\mathbb{A}$, we use a kinematic bicycle model~\cite{MATUTE2019kinematic} as our dynamic model to generate realistic trajectories that can be driven in the real world. Representing the behavior of actors as kinematic bicycle model trajectories allows for physical feasibility and fine-grained behavior control. 
To generate realistic trajectories, we first parameterize the trajectory $\mathbf{S}=\{s^t\}_0^T$ as a sequence of kinematic bicycle model states $s^t = \{p^t, \kappa^t, a^t\}$, where $p$ represents the position, $\kappa$ represents the trajectory curvature, and $a$ represents the acceleration. Then, trajectories can be generated by controlling the change of curvature $\Dot{\kappa}_t$ and the acceleration $a_t$ over time, and using the kinematic bicycle model to update corresponding other states for each timestamp. 
To generate diverse trajectories, we set the objectives as biasing the trajectories to a given direction (e.g. forward, backward, left and right), while not colliding with other agents. To that end, we optimize a carefully-designed objective function $\mathcal{L}_\text{dyn}$ over the control actions, i.e. $\Dot{\kappa}_t$ and $a_t$ for each agents. More specifically, the objective function consists of two components:
$\mathcal{L}_\text{dyn} = \mathcal{L}_\text{d} + \gamma \mathcal{L}_\text{col}$
, where $\mathcal{L}_\text{d}$ is the deviation objective loss, $\mathcal{L}_\text{col}$ is the collision regularization loss, and $\gamma$ is a weight factor to balance the objectives. In each scene, we randomly pick a deviation objective loss $\mathcal{L}_\text{d}$ from the set $\{$moving forward, backward, left, right$\}$ for each agent. More specifically, the deviation objective loss $\mathcal{L}_\text{d}$ is formulated as 
$$ \mathcal{L}_\text{d} = (\mathbf{X} - \mathbf{X}_\text{aug}) \, \mathbf{\bar{d}} \text{,} $$
where $\mathbf{X}_\text{aug}$ represents the generated trajectories by perturbing the trajectories in the dataset and $\mathbf{\bar{d}}$ represents the unit vectors for the target deviation directions in the set of $\{$moving forward, backward, left, right$\}$.
And the collision regularization loss $\mathcal{L}_\text{col}$ is formulated as
$$\mathcal{L}_{\text{col}}(\textbf{X}_{\text{aug}},\textbf{X}) = \frac{1}{n-1} \sum_{i\ne \text{aug}}^{n-1} \frac{1}{\Vert \textbf{X}_{\text{aug}} - \textbf{X}_i\Vert + 1} \,\text{,}$$

We also clip the maximum deviation of the positions so that the trajectories are constrained to be in the lane.

\subsection{MPC-based Planner}
\noindent\textbf{Planner.} In this work, to demonstrate the explicit consequences of the adversarial trajectory, we implement a simple yet effective planner that uses conformal lattice~\cite{mcnaughton2011motion} for sampling paths and model predictive control (MPC)~\cite{camacho2013model} for motion planning. We call this planner \textit{MPC-based planner}.

\noindent\textbf{Planning strategy.} In this work, we consider a closed-loop planning strategy. Though for the closed-loop planning we have to replay the ground truth trajectories of other agents, we do notice reduced collisions and driving off-road consequences compared to open-loop planning and consider the closed-loop planning fashion meaningful.

\section{Experiment and Results}

\subsection{More details on Experimental Setup}
\paragraph{Models.}
Since the adversarial training process is computationally heavy, we use a lightweight version of the AgentFormer in the analysis and ablation studies, namely mini-AgentFormer.
In mini-AgentFormer, we (1) remove the map context and (2) reduce the transformer layer from two layers to a single layer. 
We report the final results for all three models: AgentFormer (AF), mini-AgentFormer (mini-AF) and Social-GAN.

\paragraph{Evaluating impacts to downstream planners.}
To demonstrate the impacts to downstream planners, we generate adversarial examples for consecutive frames on traffic scenarios in nuScenes dataset. With the MPC-based planner plugged in, we can demonstrate the consequences of the adversarial attacks on trajectory prediction models. We use the prediction results of AgentFormer trained with different methods due to its best performance among the three models. As we mentioned in the main paper, we show 10 cases where the AV collides with other vehicles under attack. We visualize 3 scenarios in the demo video of the supplementary material.

\paragraph{Hyperparameter choices.}
To select the hyperparameter $\beta$, 
we conduct adversarial training with different $\beta$ for controlling the regularization. The results are shown in Table~\ref{tab:reg_beta_ablation}. We find that $\beta=0.1$ achieves a good trade-off between robustness and clean performance. Therefore, we use $\beta=0.1$ for the experiments in the rest of the paper.

\begin{table}[!h]
    \centering
    \begin{tabular}{c|ccccc}
    \toprule
$\beta$       & 0.01 & 0.1  & 0.5  & 1    & 10   \\\midrule
ADE        & 2.19 & 2.29 & 2.37 & 2.39 & 2.57 \\
Robust ADE & 3.91 & 3.76 & 3.80 & 3.78 & 3.79 \\\bottomrule
\end{tabular}
    \caption{Ablation study on different regularization loss weights.}
    \label{tab:reg_beta_ablation}
\end{table}

\begin{figure}[!h]
    \begin{minipage}[c]{0.45\linewidth}
    \centering
      \includegraphics[width=\textwidth]{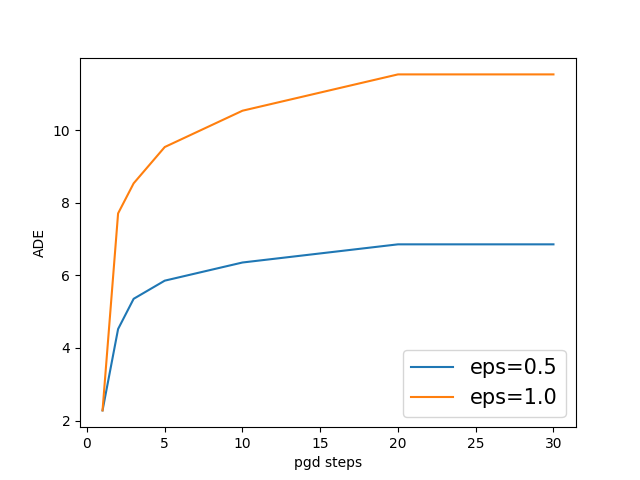}
    \caption{PGD step convergence for attack convergence with \attack. Attack converges around 20 steps.}
    \label{fig:pgd_attack_convergence}
    \end{minipage}\hfill
    \begin{minipage}[c]{0.45\linewidth}
      \centering
    \includegraphics[width=\textwidth]{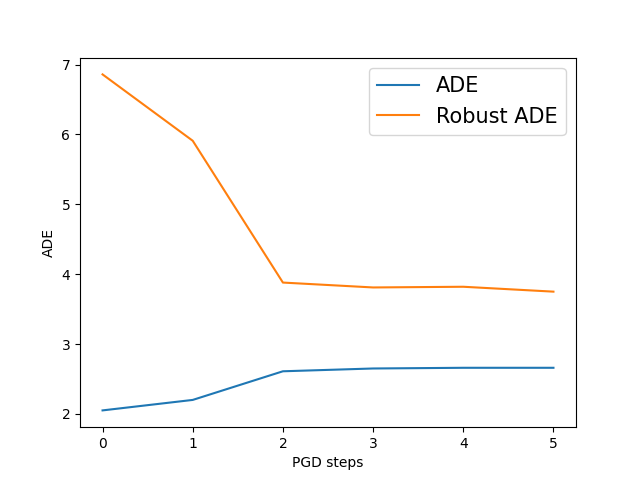}
    \caption{PGD step sizes ablation study. We find that except for 1 step PGD adversarial training, adversarial training with all the other step sizes achieves similar results. }
    \label{tab:pgd_step_ablation}
    \end{minipage}
\end{figure}

To select the PGD attack step for evaluation, we conduct ablation experiments to show the convergence of different PGD steps.
As shown in Figure~\ref{fig:pgd_attack_convergence}, the attack converges at 20 steps. Thus, we select the 20-step PDG attack for the experiments in this paper


To select the PGD attack steps of adversarial training, 
we  conduct experiments on adversarially training the model with different PGD steps. Since we are using PGD attack with adaptive step sizes~\cite{croce2020reliable}, attacks with any PGD steps are able to fully utilize the attack capability controlled by $\epsilon$. In Figure~\ref{tab:pgd_step_ablation}, we show that except for the 1-step PGD attack, all other steps show the similar robustness and clean performance.

\paragraph{Evaluation with existing attack~\cite{zhang2022adversarial}.}
We also evaluate the robust trained model with the existing attack~\cite{zhang2022adversarial}. 
The results are shown in the Table~\ref{tab:robust_traj_eval_search}. We show that the existing attack increased less prediction error (e.g., 78\% less for $\epsilon=1.0$ on AgentFormer trained with clean data) due to the additional constraints of the attack. We demonstrate that the proposed \method achieves both best robustness and least clean performance degradation, compared to the baselines.
\begin{table}[h]
\caption{Evaluation results of the proposed methods and existing methods on attacks proposed by Zhang et al.~\cite{zhang2022adversarial}. mini-AF, AF and SGAN represent mini-AgentFormer, AgentFormer, and Social-GAN respectively. \textit{DA} represents data augmentation with adversarial examples.}
\label{tab:robust_traj_eval_search}
\centering
\adjustbox{max width=\textwidth}{
\begin{tabular}{l|l|cc|cc|cc|cc}
\toprule
Model                            & Method         & \multicolumn{2}{c|}{ADE} & \multicolumn{2}{c|}{Robust ADE} & \multicolumn{2}{c|}{FDE} & \multicolumn{2}{c}{Robust FDE} \\
                                  &                & 0.5        & 1.0        & 0.5            & 1.0           & 0.5        & 1.0        & 0.5            & 1.0         \\\midrule
\multirow{5}{*}{mini-AF} & Clean                      & 2.05 & 2.05 & 3.24 & 4.61 & 4.41 & 4.41 & 6.19 & 8.02  \\
 & \textit{Na\"ive AT }                  & 2.75 & 2.78 & 3.38 & 4.46 & 5.92 & 5.89 & 6.97 & 8.17  \\
 & \textit{DA + Train-time Smoothing}  & 2.41 & 2.39 & 3.28 & 4.04 & 5.05 & 5.09 & 6.36 & 8.33  \\
 & \textit{Detection + Test Smoothing} & 2.31 & 2.28 & 3.26 & 4.41 & 4.96 & 4.91 & 6.41 & 8.27  \\
 & \textbf{\method}                    & 2.14 & 2.11 & 2.50 & 2.51 & 4.36 & 4.35 & 5.07 & 5.11  \\\midrule
\multirow{5}{*}{AF}      & Clean                      & 1.86 & 1.86 & 2.62 & 3.34 & 3.89 & 3.89 & 5.22 & 6.72  \\
      & \textit{Na\"ive AT}                   & 2.52 & 2.56 & 2.86 & 3.52 & 5.18 & 5.32 & 5.68 & 6.75  \\
      & \textit{DA + Train-time Smoothing}  & 2.17 & 2.13 & 2.72 & 3.44 & 4.59 & 4.51 & 5.38 & 6.17  \\
      & \textit{Detection + Test Smoothing} & 2.08 & 2.03 & 2.58 & 3.29 & 4.43 & 4.26 & 5.49 & 6.22  \\
      & \textbf{\method}                    & 1.91 & 1.95 & 2.14 & 2.21 & 4.02 & 4.01 & 4.31 & 4.31  \\\midrule
\multirow{5}{*}{SocialGAN}       & Clean                      & 4.80 & 4.80 & 6.45 & 8.08 & 5.52 & 5.52 & 7.78 & 11.12 \\
       & \textit{Na\"ive AT}                   & 6.43 & 6.55 & 6.99 & 8.66 & 7.60 & 7.53 & 8.98 & 10.54 \\
      & \textit{DA + Train-time Smoothing}  & 5.63 & 5.61 & 6.42 & 8.01 & 6.44 & 6.41 & 8.34 & 9.88  \\
       & \textit{Detection + Test Smoothing} & 5.35 & 5.37 & 6.34 & 8.72 & 6.12 & 6.07 & 7.77 & 10.21 \\
        & \textbf{\method}                    & 4.95 & 5.07 & 5.01 & 5.49 & 5.72 & 5.73 & 6.68 & 6.40  \\\bottomrule
\end{tabular}
}

\end{table}

\subsection{Main Results}

We evaluate our methods and existing methods with four metrics in Table~\ref{tab:main_tab_supp}. We observe that the results are consistent where the proposed \method{} achieves the best results compared to the baselines and existing methods~\cite{zhang2022adversarial}.

In the demo video, we visualized scenarios where adversarial attacks on trajectory prediction models lead to collisions on both model trained on clean data and model trained with an existing defense~\cite{zhang2022adversarial}, while model trained with \method{} is able to avoid the collisions.

\begin{table}[h]
\caption{Additional evaluation results of the proposed methods and existing methods. mini-AF, AF and SGAN represent mini-AgentFormer, AgentFormer, and Social-GAN respectively. \textit{DA} represents data augmentation with adversarial examples.}
\label{tab:main_tab_supp}
\centering
\adjustbox{max width=\textwidth}{
\begin{tabular}{l|l|cc|cc|cc|cc}
\toprule
Model                            & Method         & \multicolumn{2}{c|}{ADE} & \multicolumn{2}{c|}{Robust ADE} & \multicolumn{2}{c|}{FDE} & \multicolumn{2}{c}{Robust FDE} \\
                                  &                & 0.5        & 1.0        & 0.5            & 1.0           & 0.5        & 1.0        & 0.5            & 1.0         \\\midrule
\multirow{8}{*}{mini-AF} & Clean          & 2.05       & 2.05       & 6.86           & 11.53         & 4.41       & 4.41       & 13.08          & 20.15     \\\cmidrule{2-10}
                                  & \textit{Na\"ive AT }      & 2.75       & 2.78       & 5.44           & 9.20          & 5.92       & 5.89       & 10.13          & 15.78             \\
                                  & \textit{DA }      & 2.31 & 2.32 & 5.54 & 9.32 & 5.01 & 4.92 & 10.09 & 15.77            \\
                                 & \textit{Train-time Smoothing }     & 3.14 & 3.07 & 5.67 & 9.31 & 6.77 & 6.61 & 10.51 & 17.48         \\
                                & \textit{Test-time Smoothing}      & 2.97 & 3.07 & 4.96 & 8.50 & 6.49 & 6.31 & 9.25  & 14.13             \\
                                  & \textit{DA + Train-time Smoothing}      & 2.41       & 2.39       & 5.48           & 9.00          & 5.05       & 5.09       & 10.23          & 16.87               \\
                                  & \textit{Detection + Test Smoothing} & 2.31       & 2.28       & 5.91           & 9.85          & 4.96       & 4.91       & 11.49          & 17.57          \\
                                  & \textbf{\method{} }       & \textbf{2.14}       & \textbf{2.11}       & \textbf{3.69}           & \textbf{3.82}          & \textbf{4.36}       & \textbf{4.35}       & \textbf{7.10}           & \textbf{7.59}              \\\midrule
\multirow{8}{*}{AF}      & Clean          & 1.86       & 1.86       & 5.09           & 8.57          & 3.89       & 3.89       & 9.42           & 14.41              \\\cmidrule{2-10}
                                  & \textit{Na\"ive AT }      & 2.52       & 2.56       & 3.81           & 6.81          & 5.18       & 5.32       & 7.11           & 10.76              \\
                                  & \textit{DA }      & 2.10 & 2.08 & 4.35 & 7.22 & 4.33 & 4.38 & 8.08 & 12.15           \\
                                 & \textit{Train-time Smoothing }     & 2.11 & 2.13 & 4.19 & 6.79 & 4.40 & 4.46 & 8.01 & 11.13         \\
                                & \textit{Test-time Smoothing}      & 2.40 & 2.41 & 4.43 & 7.44 & 5.02 & 4.99 & 8.23 & 12.47             \\
                                  & \textit{DA + Train-time Smoothing}      & 2.17       & 2.13       & 4.14           & 6.62          & 4.59       & 4.51       & 7.85           & 11.00          \\
                                  & \textit{Detection + Test Smoothing} & 2.08       & 2.03       & 4.45           & 7.59          & 4.43       & 4.26       & 8.01           & 12.74             \\
                                  & \textbf{\method{}}        & \textbf{1.91}       & \textbf{1.95}       & \textbf{2.73}           & \textbf{2.86}          & \textbf{4.02}       & \textbf{4.01 }      & \textbf{5.22}           & \textbf{5.48}              \\\midrule
\multirow{8}{*}{SGAN}        & Clean          & 4.80       & 4.80       & 10.52          & 20.15         & 5.52       & 5.52       & 15.60          & 24.79              \\\cmidrule{2-10}
                                  & \textit{Na\"ive AT}       & 6.43       & 6.55       & 8.34           & 14.63         & 7.60       & 7.53       & 13.71          & 17.93          \\
                                  & \textit{DA }      & 5.41 & 5.40 & 8.85 & 17.25 & 6.16 & 6.21 & 13.33 & 20.83           \\
                                 & \textit{Train-time Smoothing }     & 5.50 & 5.47 & 8.74 & 16.51 & 6.27 & 6.31 & 14.03 & 19.48         \\
                                & \textit{Test-time Smoothing}      & 6.16 & 6.17 & 9.05 & 17.42 & 7.14 & 7.07 & 13.52 & 21.81             \\
                                  & \textit{DA + Train-time Smoothing}      & 5.63       & 5.61       & 8.60           & 16.14         & 6.44       & 6.41       & 13.82          & 19.08           \\
                                  & \textit{Detection + Test Smoothing} & 5.35       & 5.37       & 9.28           & 17.39         & 6.12       & 6.07       & 13.36          & 21.59            \\
                                  & \textbf{\method{}}        & \textbf{4.95}       & \textbf{5.07}      & \textbf{5.20}           & \textbf{6.94}          & \textbf{5.72}       & \textbf{5.73}       & \textbf{8.97}           & \textbf{8.89}           \\\midrule
Model                      & Method                          & \multicolumn{2}{c|}{MR} & \multicolumn{2}{c|}{Robust MR} & \multicolumn{2}{c|}{ORR} & \multicolumn{2}{c}{Robust ORR} \\
                           &                                 & 0.5  & 1.0  & 0.5       & 1.0  & 0.5  & 1.0  & 0.5        & 1.0  \\
                           \midrule
\multirow{8}{*}{mini-AF}   & Clean                           & 0.33 & 0.33 & 0.77      & 0.93 & 0.08 & 0.08 & 0.28       & 0.45 \\\cmidrule{2-10}
                           & \textit{Na\"ive AT}                        & 0.45 & 0.45 & 0.60      & 0.70 & 0.11 & 0.10 & 0.22       & 0.34 \\
                           & \textit{DA }                             & 0.37 & 0.38 & 0.61      & 0.72 & 0.09 & 0.09 & 0.22       & 0.35 \\
                           & \textit{Train-time Smoothing}            & 0.50 & 0.50 & 0.66      & 0.79 & 0.12 & 0.12 & 0.22       & 0.38 \\
                           & \textit{Test-time Smoothing}             & 0.48 & 0.50 & 0.56      & 0.65 & 0.11 & 0.11 & 0.20       & 0.32 \\
                           & \textit{DA + Train-time Smoothing}       & 0.39 & 0.40 & 0.65      & 0.77 & 0.09 & 0.09 & 0.22       & 0.37 \\
                           & \textit{Detection + Test-time Smoothing} & 0.37 & 0.38 & 0.69      & 0.81 & 0.09 & 0.09 & 0.25       & 0.41 \\
                           &\textbf{ \method{} }                     & \textbf{0.34} & \textbf{0.36} & \textbf{0.54}      & \textbf{0.54} & \textbf{0.08} & \textbf{0.08} & \textbf{0.10}       & \textbf{0.11} \\\midrule
\multirow{8}{*}{AF}        & Clean                           & 0.29 & 0.29 & 0.66      & 0.88 & 0.04 & 0.04 & 0.16       & 0.30 \\\cmidrule{2-10}
                           & \textit{Na\"ive AT}                        & 0.39 & 0.38 & 0.51      & 0.69 & 0.06 & 0.06 & 0.13       & 0.22 \\
                           & \textit{DA }                             & 0.32 & 0.32 & 0.56      & 0.74 & 0.05 & 0.05 & 0.14       & 0.26 \\
                           & \textit{Train-time Smoothing}            & 0.33 & 0.33 & 0.56      & 0.71 & 0.05 & 0.05 & 0.13       & 0.25 \\
                           & \textit{Test-time Smoothing}             & 0.37 & 0.37 & 0.58      & 0.77 & 0.05 & 0.05 & 0.14       & 0.26 \\
                           & \textit{DA + Train-time Smoothing}       & 0.33 & 0.33 & 0.54      & 0.70 & 0.05 & 0.05 & 0.13       & 0.24 \\
                           & \textit{Detection + Test-time Smoothing} & 0.32 & 0.33 & 0.59      & 0.76 & 0.04 & 0.05 & 0.14       & 0.27 \\
                           & \textbf{\method{}}                      & \textbf{0.29} & \textbf{0.31} & \textbf{0.46}      & \textbf{0.51} & \textbf{0.04} & \textbf{0.04} & \textbf{0.05}       & \textbf{0.07} \\\midrule
\multirow{8}{*}{SocialGAN} & Clean                           & 0.40 & 0.40 & 0.85      & 0.99 & 0.14 & 0.14 & 0.52       & 0.60 \\\cmidrule{2-10}
                           & \textit{Na\"ive AT}                        & 0.53 & 0.53 & 0.63      & 0.77 & 0.19 & 0.19 & 0.39       & 0.44 \\
                           & \textit{DA}                              & 0.44 & 0.44 & 0.72      & 0.85 & 0.16 & 0.16 & 0.44       & 0.51 \\
                           & \textit{Train-time Smoothing }           & 0.45 & 0.45 & 0.67      & 0.82 & 0.16 & 0.16 & 0.42       & 0.50 \\
                           & \textit{Test-time Smoothing}             & 0.51 & 0.51 & 0.74      & 0.85 & 0.19 & 0.19 & 0.45       & 0.52 \\
                           & \textit{DA + Train-time Smoothing}       & 0.47 & 0.46 & 0.66      & 0.81 & 0.17 & 0.17 & 0.41       & 0.49 \\
                           & \textit{Detection + Test-time Smoothing} & 0.45 & 0.44 & 0.74      & 0.89 & 0.16 & 0.16 & 0.46       & 0.53 \\
                           & \textbf{\method{}}                      & \textbf{0.41} & \textbf{0.42} & \textbf{0.60}      & \textbf{0.62} & \textbf{0.15} & \textbf{0.15} & \textbf{0.24}       & \textbf{0.29} \\
\bottomrule
\end{tabular}}
\end{table}

\end{document}